\documentclass[10pt,twocolumn,letterpaper]{article}
\usepackage[table]{xcolor}

\usepackage{cvpr}      
\usepackage{multirow}
\definecolor{cvprblue}{rgb}{0.21,0.49,0.74}
\usepackage[pagebackref,breaklinks,colorlinks,allcolors=cvprblue]{hyperref}

\def\paperID{*****} 
\def\confName{CVPR}
\def\confYear{2025}

\title{Leveraging Pre-Trained Visual Models for AI-Generated Video Detection}

\author{Keerthi Veeramachaneni \\
Georgia Institute of Technology \\
{\tt\small aveerama3@gatech.edu }
\and
Praveen Tirupattur\\
Univeristy of Central Florida \\
{\tt\small praveen.tirupattur@ucf.edu}
\and
Amrit Singh Bedi\\
Univeristy of Central Florida \\
{\tt\small amritbedi@ucf.edu}
\and
Mubarak Shah\\
Univeristy of Central Florida \\
{\tt\small shah@crcv.ucf.edu}
}

\begin{document}
\maketitle
\begin{abstract}

    Recent advances in Generative AI (GenAI) have led to significant improvements in the quality of generated visual content. As AI-generated visual content becomes increasingly indistinguishable from real content, the challenge of detecting the generated content becomes critical in combating misinformation, ensuring privacy, and preventing security threats. Although there has been substantial progress in detecting AI-generated images, current methods for video detection are largely focused on deepfakes, which primarily involve human faces. However, the field of video generation has advanced beyond DeepFakes, creating an urgent need for methods capable of detecting AI-generated videos with generic content. To address this gap, we propose a novel approach that leverages pre-trained visual models to distinguish between real and generated videos. The features extracted from these pre-trained models, which have been trained on extensive real visual content, contain inherent signals that can help distinguish real from generated videos. Using these extracted features, we achieve high detection performance without requiring additional model training, and we further improve performance by training a simple linear classification layer on top of the extracted features. We validated our method on a dataset we compiled (VID-AID), which includes around 10,000 AI-generated videos produced by 9 different text-to-video models, along with 4,000 real videos, totaling over 7 hours of video content. Our evaluation shows that our approach achieves high detection accuracy, above 90\% on average, underscoring its effectiveness. Upon acceptance, we plan to publicly release the code, the pre-trained models, and our dataset to support ongoing research in this critical area.
   
\end{abstract}   
\section{Introduction}
%

\begin{figure}[h!]
  \centering
  \includegraphics[width=0.93\linewidth]{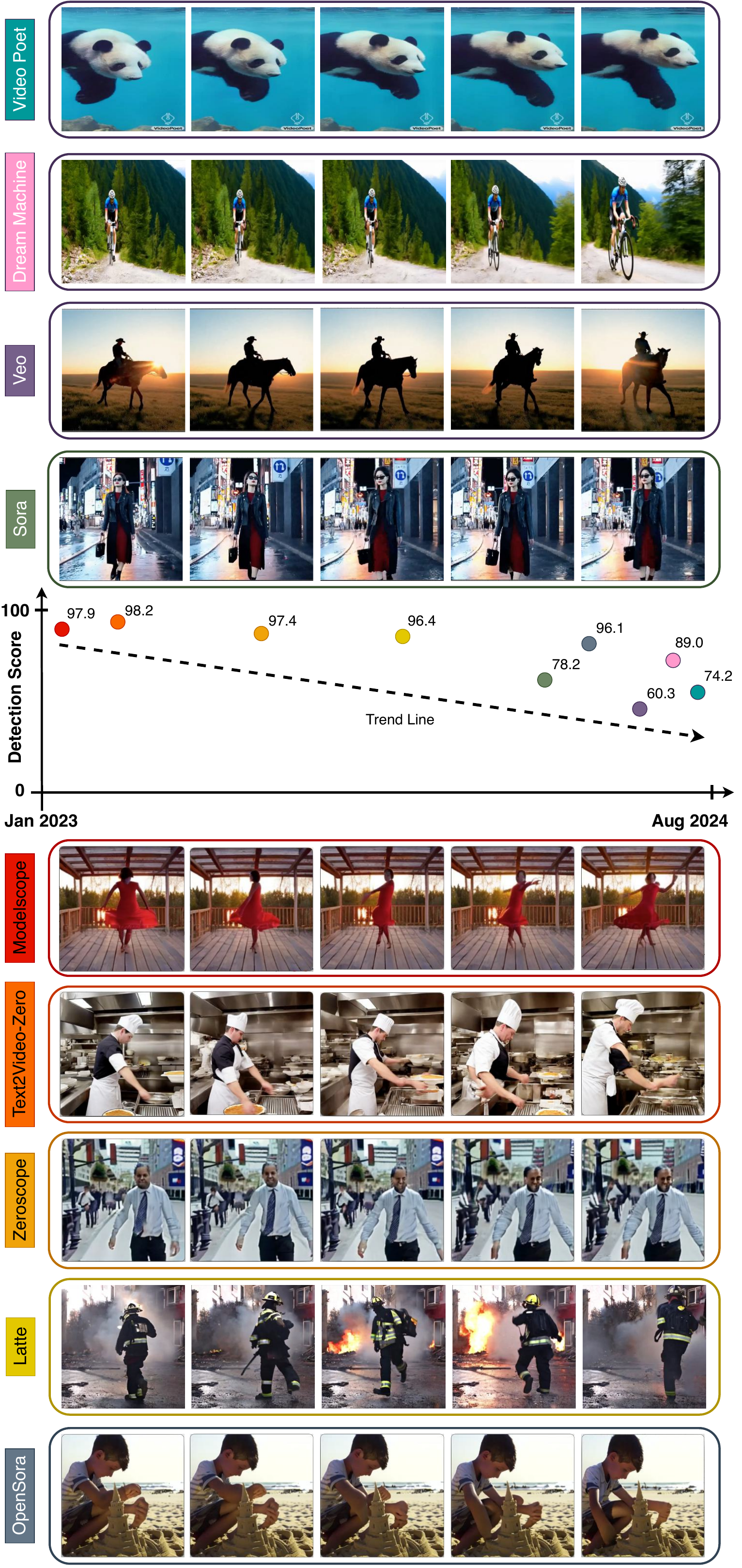}
    \caption{Timeline illustrating the advancements in text-to-video generation models, accompanied by example frames from videos generated by multiple text-to-video (T2V) models. \textbf{Top:} Sample frames from closed-source models, such as Sora \cite{videoworldsimulators2024} and Veo \cite{veo}, that generate high-quality videos with better spatial and temporal consistencies. \textbf{Middle:} Timeline depicting the rapid progress in the development of T2V models. The quality of the generated videos has increased drastically, making them indistinguishable from real videos and making them harder to detect. \textbf{Bottom:} Sample frames from videos generated by different open-source text-to-video models, showing temporal inconsistencies and spatial artifacts, such as patch-level irregularities. For the best experience, view in color and zoom in to observe the visual artifacts.}
   \label{fig:text-to-video}
\end{figure}

In recent years, the digital landscape has witnessed a significant surge in AI-generated visual content, fueled by rapid advancements in Generative AI (GenAI). Although the initial focus of these technologies was primarily on image creation, recent developments have expanded GenAI's capabilities to produce highly realistic video content. This evolution from static images to dynamic videos marks a new frontier in content creation using artificial intelligence, driven by technological progress and the increasing demand for more immersive media experiences.

The primary motivation behind developing generative models is to minimize manual effort while enhancing creative possibilities. These systems offer content creators unprecedented opportunities by automating video production, saving time and resources. They also enable the creation of innovative and imaginative content that would be difficult or impossible to achieve through traditional methods. However, along with these advantages comes a significant risk of misuse. Generative models can be exploited to create realistic yet deceptive videos, particularly in the realms of politics and economics. For example, a generative model could produce a convincing but fake news broadcast or political speech, potentially misleading the public and influencing opinions or decisions based on false information. Given the potential for such harmful misuse, there is a pressing need to develop robust methods to detect AI-generated videos.

Although Deepfake detection \cite{zhao2021multi, cozzolino2023audio, yan2023ucf} has been extensively researched, these methods are mainly focused on videos featuring people, particularly when faces are fully visible. They concentrate on identifying manipulations in facial expressions and movements but are less effective for detecting other types of AI-generated videos, especially those without human faces or where faces are obscured. We propose an approach to detect generic AI-generated videos that seek to overcome the limitations of current methods by providing a more comprehensive solution to address the challenges posed by AI-generated videos.

\vspace{0.1cm}
In this work, our goal is to advance the development of robust methods for detecting AI-generated videos, with a particular focus on those produced by text-to-video models. To facilitate the creation of reliable detection systems, we have compiled a dataset of generated videos from nine different text-to-video generative models, including five open-source and four closed-source models. This dataset serves as a crucial resource for training and evaluating detection methods. In addition to collecting this dataset, we introduce a novel solution for detecting AI-generated videos.

\vspace{0.1cm}
Our approach leverages pre-trained visual models that are trained on a large corpus of real videos to detect generated videos. We hypothesize that features from these pre-trained models are rich in the information required to differentiate between videos from the real world and AI-generated videos. By detecting AI-generated videos, our approach aims to combat the spread of misinformation and enhance the reliability of digital media. 

The main contributions of our work include: 
\begin{itemize}
  \item We propose a novel approach for detecting AI-generated video content by leveraging features from pre-trained large vision models to distinguish between AI-generated and real videos. This method eliminates the need to collect large datasets and train extensive models, offering an efficient solution to the problem.
  
  \item We compiled a dataset of AI-generated videos, named VID-AID, with videos from nine text-to-video generative models, including five open-source and four closed-source models, to support the development and evaluation of detection methods. Our dataset consists of more than 7 hours of video content, making it a valuable resource for developing and evaluating detection methods for AI-generated content.
   
  \item We evaluate our proposed method on the collected dataset and provide an extensive analysis of our results. We propose two evaluation protocols to check the generalization capability of our approach and show that our approach achieves high detection performance, over $90\%$ on average, on both open-source and closed-source text-to-video models.
\end{itemize}

\section{Related Work}
\subsection{Visual Content Generation}

Video generation methods have evolved into powerful tools for producing high-quality video content from textual (T2V) or image (I2V) prompts \cite{yangcross, hu2024animate, wu2023tune}. Early T2V approaches, like LVDM \cite{Blattmann_2023_CVPR} and ModelScope \cite{wang2023modelscope}, modified 2D image diffusion models by transforming the U-Net architecture into a 3D U-Net and training it on extensive video datasets. Building on this foundation, methods such as AnimatedDiff \cite{guoanimatediff} integrated temporal attention modules into existing 2D latent diffusion models, preserving the strong performance of T2I models. More recently, transformer-based diffusion techniques \cite{bar2024lumiere, gupta2023photorealistic, ma2024latte} have enabled large-scale, joint training across both videos and images, leading to notable advancements in generation quality. Most open-source video generation models \cite{wang2023modelscope, ma2024latte, opensora} produce short videos, typically 16 frames at 8 fps. Recent methods have explored long video generation, aiming to create videos lasting a few minutes with holistic visual consistency \cite{villegas2022phenaki, he2023animate, yin2023nuwa}. These longer videos often exhibit repetitive patterns of a single action without transitions. However, recently released closed-source models \cite{kondratyukvideopoet, videoworldsimulators2024, veo, dreammachine} exhibit the ability to generate longer videos, up to a few minutes, of more dynamic scenes with transitions. In this work, we mainly focus on detecting videos generated by T2V models. 

\subsection{AI-Generated Content Detection}
Research in AI-generated content detection has predominantly concentrated on images rather than videos. In the realm of fake image detection, approaches can be broadly categorized into training-free and training-based methods.  \citet{yang2024d3} and \citet{Wang2023DIRE} explore universal artifacts and the reconstruction of fake and real images, respectively. \citet{wang2020cnn} employs a classification technique, demonstrating that a model trained on one generator can generalize to other generators of the same type, such as different variants of GANs. Building on this, \citet{ojha2023towards} extends the focus towards achieving generalizability and universal fake detection by leveraging a feature space that was not explicitly trained for this purpose.

The challenges existing detection methods face include finding artifacts shared by different generation models while keeping computational costs relatively low. Finding a comprehensive dataset for training also remains a challenge, especially for the task of AI-generated video detection. Although there are a few open-source generation models, the videos collected from these models can differ in quality from state-of-the-art models. This poses a problem as detection models trained using these videos might not be able to generalize to new generation models that generate higher quality videos, making it harder for detection models. 

The field of generated video content detection is still in its infancy, with current research primarily targeting Deepfake detection rather than the broader spectrum of generated video content. Approaches in this area can similarly be classified into training-based methods and training-free techniques. \citet{guera2018deepfake} proposed a method that uses a convolutional network to extract features from Deepfake videos, which are then processed by an RNN to determine whether the video has been manipulated. Unlike fake images, fake videos present unique challenges, as they can exhibit both spatio-temporal inconsistencies and detectable biological signals. Recently, novel approaches such as gaze inconsistency analysis, as explored by \citet{peng2024deepfakes}, have gained traction. However, these techniques often struggle with other types of AI-generated videos that may lack such signals. For instance, many AI-generated videos do not feature humans or contain humans with their faces not facing the camera, making methods that rely on biological indicators \cite{mao2021deepfake}, ineffective.

In this work, we aim to address the challenge of detecting AI-generated videos with diverse visual content. To tackle this problem, we propose both a training-free method and a trained approach that leverages an efficient classification model requiring minimal data. Additionally, we compile a dataset for evaluating our methods, which includes videos from nine different text-to-video (T2V) models, encompassing both open-source options and advanced models like Veo \cite{veo}, Sora \cite{videoworldsimulators2024}, and the newly released Dream Machine \cite{dreammachine} and VideoPoet \cite{kondratyukvideopoet}.

\section{Dataset}
To advance the development and evaluation of detection models, it is imperative to have a dataset comprising video samples generated by T2V models. Such a dataset should include diverse samples from a variety of T2V models to rigorously test the detection model’s generalizability across different generation techniques. Furthermore, it should feature high-quality and realistic videos generated to ensure a robust evaluation, as the quality of the generated content plays a crucial role in influencing the evaluation results. In response to this critical requirement, we have curated a video dataset (VID-AID) of generated content, distinguished by three primary characteristics:

\paragraph{Diverse Content}
The availability of open-source T2V models allows us to generate multiple videos using different inputs as text. To create a dataset with diverse videos, we employ GPT-3.5 \cite{openai2023chatgpt} to generate 1K captions using the following prompt:

\textit{
Can you generate sample captions for a 2-3 second long video? Each caption should describe the scene, the subjects, the objects, and the actions being performed in the video.}

The generated captions depict various scenes with different subjects, both human and non-human, engaging in a range of actions. This diversity in captions provides the variety needed to create videos with diverse content for our dataset. We provide these captions used to generate the videos as part of our dataset.

\paragraph{High Quality} Recent models like Sora \cite{videoworldsimulators2024} and Veo \cite{veo} produce high-quality videos, a marked improvement over previous open-source models \cite{wang2023modelscope, ma2024latte, opensora}. However, these models are not publicly accessible, creating a significant challenge in collecting a high-quality dataset. We mitigate this by gathering all the available videos generated by the closed-source models posted on various social media platforms, ensuring that our dataset has many high-quality samples from the latest text-to-video (T2V) generation models.

\paragraph{Multiple T2V Models} To thoroughly assess the effectiveness of our proposed approach, it is crucial to evaluate the detection models using videos generated by multiple T2V models. Accordingly, our dataset includes videos from nine different T2V generation models, comprising five open-source models and four closed-source models. For detailed information on the T2V models and the corresponding video statistics in our dataset, please refer to Table \ref{tab:dataset}. To ensure a balanced dataset for comparative analysis, we have also included real videos sourced from the YouTube-VOS dataset \cite{xu2018youtube}, alongside the generated videos from the T2V models.

The videos generated by the open-source T2V models are each 2 seconds in length, while those from the latest closed-source models vary in duration, with some extending up to a minute. To maintain consistency across the dataset, we split all videos longer than 2 seconds into non-overlapping 2-second clips, treating each clip as a distinct video instance. Following this pre-processing step, our dataset comprises a total of 14,000 videos, including 10,000 videos from nine different T2V generation models and 4,000 real videos from the YouTube-VOS dataset \cite{xu2018youtube}. This extensive and diverse collection enables a thorough and comprehensive evaluation of our approach, ensuring that it is rigorously tested against a wide range of high-quality video content.

\begin{table*}[h!]
\centering
\small
\begin{tabular}{l|c|c|c|c|c|c}
\hline
\textbf{Source} & \textbf{Type} & \textbf{Duration} & \textbf{Resolution} & \textbf{FPS} & \textbf{Video Length} & \textbf{Count} \\ 
 &  & \textbf{(min.)} &  &  & \textbf{(sec.)} &  \\ 
\hline
YouTube-VOS \cite{xu2018youtube} & Real & 133.5 & - & \{24, 30\} & 2 & 4005 \\ \hline
ModelScope \cite{wang2023modelscope} & Fake & 33.3 & 256$\times$256 & 8 & 2 & 1000 \\  
Text2Video-Zero \cite{khachatryan2023text2video} & Fake & 33.3 & 512$\times$512 & 8 & 2 & 1000 \\ 
Zeroscope \cite{zeroscope} & Fake & 33.3 & 256$\times$256 & 8 & 2 & 1000 \\ 
Latte \cite{ma2024latte} & Fake & 33.3 & 512$\times$512 & 8 & 2 & 1000 \\ 
OpenSora \cite{opensora} & Fake & 33.3 & 512$\times$512 & 8 & 2 & 1000 \\ \hline
Sora \cite{videoworldsimulators2024} & Fake & 132.9 & - & \{24, 25, 30\} & 2 & 3988 \\ 
Veo \cite{veo} & Fake & 7.9 & - & \{24, 30\} & 2 & 238 \\ 
DreamMachine \cite{dreammachine} & Fake & 21.0 & - & \{24, 30\} & 2 & 631 \\ 
Video Poet \cite{kondratyukvideopoet} & Fake & 7.3 & - & 8 & 2 & 272 \\ \hline
\textbf{Total} & - & \textbf{440.8} & - & - & - & \textbf{14099} \\ \hline
\end{tabular}
\caption{Summary of our VID-AID datasets with the source of the videos and the statistics including total duration,  resolutions, FPS, video length, and the number of videos.}
\label{tab:dataset}
\vspace{-0.4cm}
\end{table*}
\section{Method}
\subsection{AI-Generated Video Detection}
Given a video, the objective is to determine whether the video is authentic or AI-generated. Our approach addresses this challenge by leveraging visual models pre-trained on a large corpus of real videos. The rationale is that these models, having learned the distribution of real videos from large-scale training, encode in their features the signal necessary to distinguish between real and AI-generated content. Using these features, in this work, we investigate training-free and training-based methods to solve the detection task. The details of these approaches, including feature extraction, are discussed below.

\paragraph{Feature Extraction} To encode both real and AI-generated videos, we employ SigLIP \cite{zhai2023sigmoid}, a pre-trained visual-language image model, and VideoMAE \cite{tong2022videomae}, a video model trained using masked modeling within a self-supervised learning framework. SigLIP \cite{zhai2023sigmoid} is an adaptation of CLIP \cite{radford2021learning}, trained on a large dataset of image-text pairs with a Sigmoid loss function. VideoMAE \cite{tong2022videomae} processes masked video inputs to reconstruct the occluded regions. We utilize the encoder from VideoMAE and the image encoder from SigLIP to extract feature representations for all videos in our dataset. When using the image-level SigLIP model, we represent the video feature as the average of the features from all the frames in the video.

\begin{figure}[h]
  \centering
  \includegraphics[width=1\linewidth]{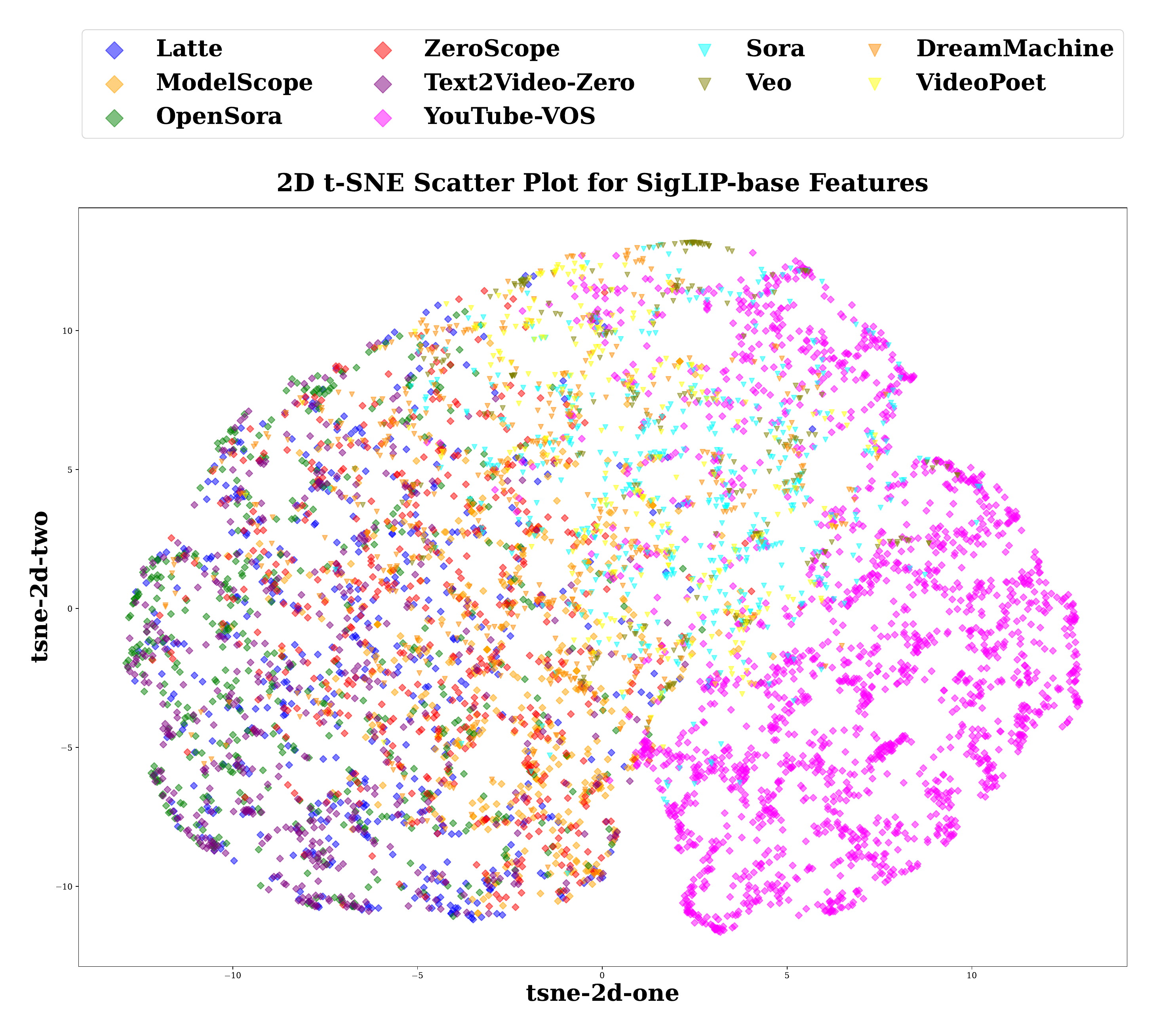}
   \caption{t-SNE visualization of SigLIP \cite{zhai2023sigmoid} features of the real videos and the videos from different T2V models in our dataset.}
   \label{fig:t-sne_plot}
   \vspace{-0.4cm}
\end{figure}

In Figure \ref{fig:t-sne_plot}, using t-SNE \cite{JMLR:v9:vandermaaten08a}, we visualize the SigLIP features for the videos in our dataset, both the real videos and the videos generated by different T2V models. From this visualization, we observe that the features corresponding to the videos generated by the open-source models and the features corresponding to the real videos group into two distinct clusters; whereas the features corresponding to the videos generated by the latest closed-source models like Sora \cite{videoworldsimulators2024} and Veo \cite{veo}, overlap with the real videos. This suggests that it is easier to detect videos generated by open-source models using these features compared to those generated by recent closed-source models.

\subsection{Training-Free Approach}
We first address the detection problem using a training-free, distance-based approach. Originally proposed by \citet{ojha2023towards} to detect generated images, this method relies on features extracted from a pre-trained CLIP \cite{radford2021learning} model. We extend this approach to detect generated videos and use features from SigLIP \cite{zhai2023sigmoid} or VideoMAE \cite{tong2022videomae} models as mentioned above.

\begin{figure}[h!]
  \centering
  \includegraphics[width=\linewidth]{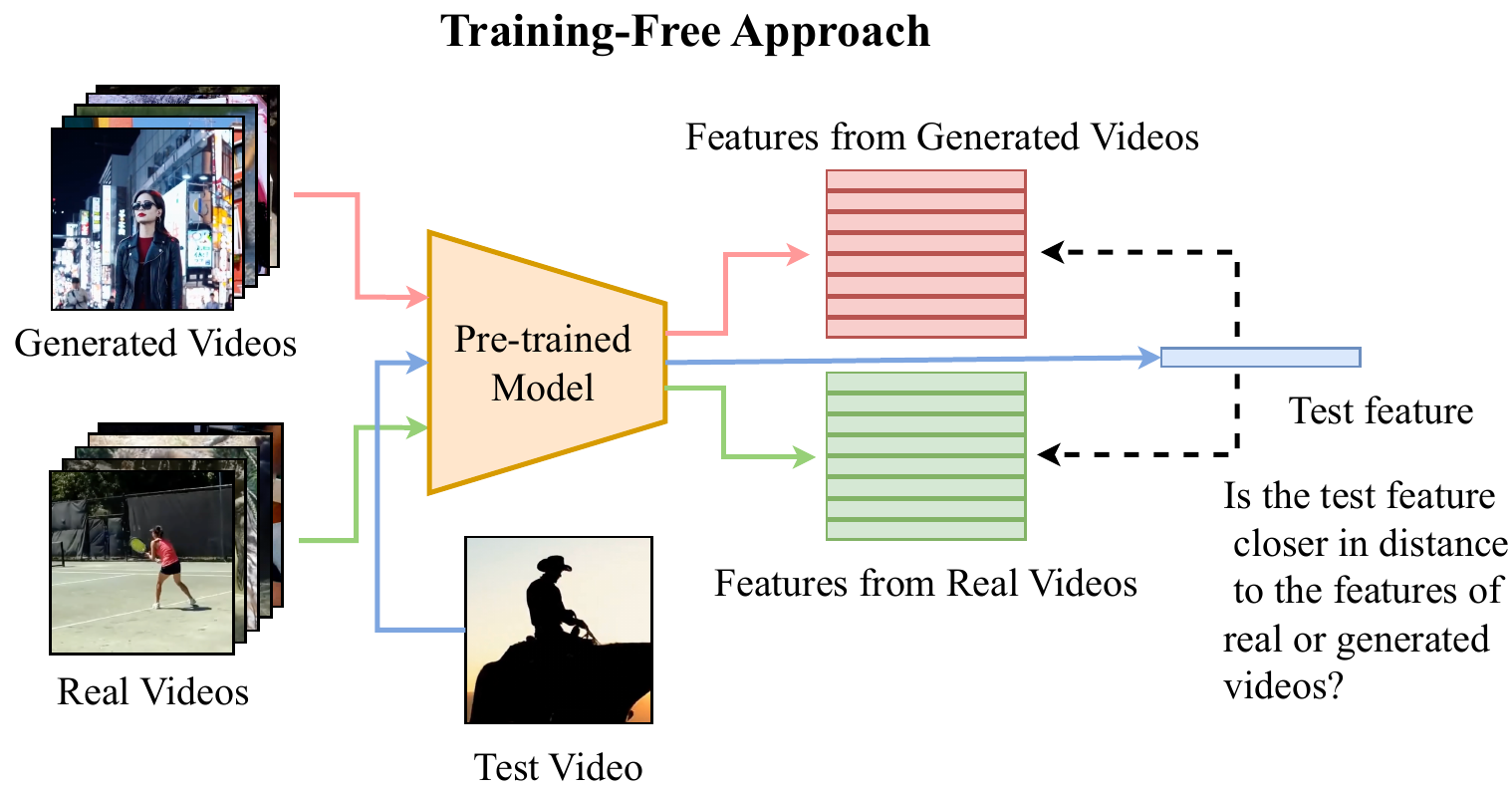}
   \caption{Overview of the training-free, distance-based approach to detect generated videos.}
   \label{fig:train_free}
   \vspace{-0.4cm}
\end{figure}

In this distance-based approach, we utilize reference sets consisting of real and generated videos. For a given test video, we compute the distance between its feature representation and those of the real and generated reference videos. The classification is then determined based on proximity: if the test video’s features are closer to those of the real videos in the reference set, it is classified as real; otherwise, it is labeled as generated. Please refer to Figure \ref{fig:train_free} for an overview of this approach. In the experiments with this approach, we consider a subset of generated videos and a subset of real YouTube-VOS videos from our dataset as the reference videos and use the others for testing. 

\subsection{Training-Based Approach}
\begin{figure}[h]
  \centering
  \includegraphics[width=1\linewidth]{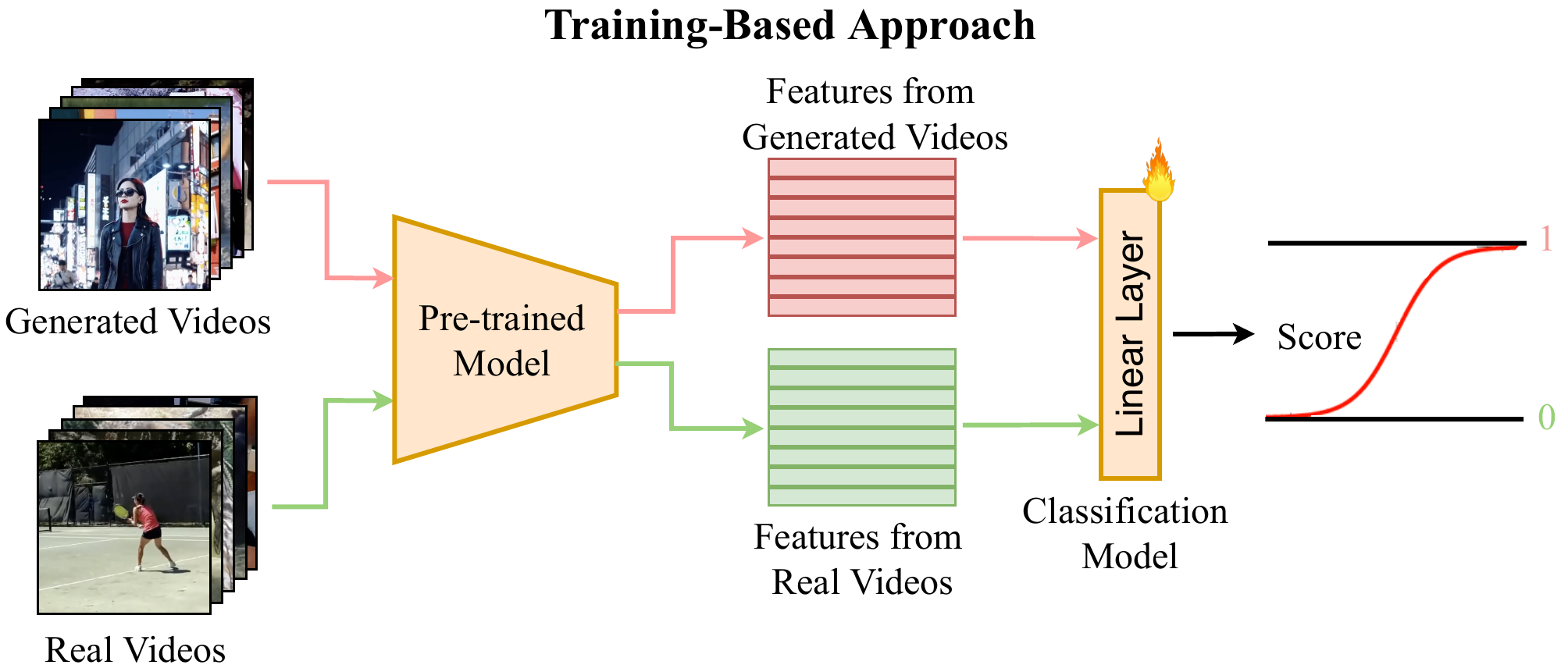}
   \caption{Overview of the training-based approach, which involves training a classification model, with single linear layers, on features extracted from pre-trained models.}
   \label{fig:train}
   \vspace{-0.4cm}
\end{figure}

The effectiveness of the training-free approach is influenced by the quality of the videos in the reference set. To overcome this limitation and improve detection performance, we also propose a training-based approach. In this method, we train a parameter-efficient linear classification model on features extracted from real and generated videos. During inference, the trained binary classifier predicts whether the input video is real or AI-generated. Figure \ref{fig:train} provides an overview of this approach. For details on the training and test splits used in our experiments, please refer to the section on the implementation details.

\section{Experiments}
\subsection{Evaluation Protocols}

Given the rapid advancement in generation methods, it is essential for detection models to generalize effectively across different generation methods. To evaluate this generalization capability, we propose cross-dataset evaluation protocols that assess the robustness of detection methods. Specifically, we introduce two protocols: 1) One-to-many generalization and 2) Many-to-many generalization.

\noindent \textbf{One-to-many generalization} In this protocol, the detection model is trained on a set of real videos and the generated videos from a single T2V model. The trained model is evaluated on a different set of real videos and the generated videos from the other T2V models. 

\noindent \textbf{Many-to-many generalization} In this protocol, the detection model is trained on a set of real videos and a set of generated videos from all the T2V models in our dataset. The trained model is then evaluated on a different set of real videos and the generated videos from all the T2V models.

\subsection{Implementation Details} \label{sec:implementation}

\paragraph{Datasets}

In the Many-to-many generalization protocol, for both the training-free (distance) approach and training-based approach, our existing dataset was split. For all the open-source models and Sora, the dataset was split into half, making sure to prevent overlap in video content, creating training and testing sets. The training set containing all of the open-source videos consists of 2500 videos, while the set with Sora videos consists of 1994 videos. Combined, the training set consists of around 4500 videos. The other closed-source models had a limited amount of videos and were only used for testing in these experiments. In the one-to-many generalization experiments, there was no dataset split, and all features from each model were used in different combinations of reference/training and testing.

\paragraph{Model Training} 

In the training-free approach, the Euclidean distance between features is used to classify the test video by using the label of the reference features with the minimum Euclidean distance to the test feature. In the training-based approach, our architecture consists of a single linear layer trained on features extracted using the pre-trained model. PyTorch \cite{paszke2019pytorch} is used for the implementation. We train our models for 100 epochs with a batch size of 32 using the Adam \cite{kingma2014adam} optimizer, with a learning rate of 1e-4. The cross-entropy loss is used to train the model and update the parameters.

\paragraph{Evaluation Metrics} To evaluate our proposed approaches, we use the F1-Score as the primary metric. This metric is calculated for both classes—real and generated. We opted for the F1-Score over accuracy because it accounts for both precision and recall, providing a more reliable measure of the detection model's performance, especially in the presence of a class imbalance in the test set.

\subsection{Results}
In this section, we present the results for both the training-free and training-based approaches with SigLIP \cite{zhai2023sigmoid} and VideoMAE \cite{tong2022videomae} features, evaluated using both the one-to-many and many-to-many protocols. Additional results with more pre-trained visual models are provided in the supplementary. In Table \ref{tab:distance_one_to_many} and Table \ref{tab:distance_many_to_many}, we show the results of our training-free approach using the one-to-many and many-to-many evaluation protocols respectively. Similarly, in Table \ref{tab:train_one_to_many} and Table \ref{tab:train_many_to_many}, we show the results of our training-based approach using the one-to-many and many-to-many evaluation protocols, respectively.

\begin{table*}[h!]
\resizebox{\linewidth}{!}{
\begin{tabular}{cccccccccccc}
\hline

\multirow{2}{*}{Reference} & \multirow{2}{*}{Model} & \multirow{2}{*}{Metric} & \multicolumn{8}{c}{Testing (real and fake)} \\ \cline{4-12}
& & & Latte & ModelScope & OpenSora & ZeroScope & Text2Video & Veo & Sora & Dream Machine & Video Poet\\ \hline

\multirow{2}{*}{Latte} & SigLIP-base & \begin{tabular}[c]{@{}c@{}}F1-Real\\ F1-Fake\end{tabular} & \begin{tabular}[c]{@{}c@{}} - \\ - \end{tabular} & \begin{tabular}[c]{@{}c@{}}99.1 \\ 98.2\end{tabular} & \begin{tabular}[c]{@{}c@{}}99.9 \\ 99.7\end{tabular} & \begin{tabular}[c]{@{}c@{}}99.1 \\ 98.2\end{tabular} & \begin{tabular}[c]{@{}c@{}}99.8 \\ 99.6\end{tabular} & \begin{tabular}[c]{@{}c@{}}95.5 \\ 35.3\end{tabular} & \begin{tabular}[c]{@{}c@{}}60.1 \\ 50.1 \end{tabular} & \begin{tabular}[c]{@{}c@{}}98.0 \\ 93.1 \end{tabular} & \begin{tabular}[c]{@{}c@{}}97.3 \\ 74.3\end{tabular} \\ \cline{2-12} 
& VideoMAE & \begin{tabular}[c]{@{}c@{}}F1-Real\\ F1-Fake\end{tabular} & \begin{tabular}[c]{@{}c@{}} - \\ -\end{tabular} & \begin{tabular}[c]{@{}c@{}}97.7 \\ 95.0\end{tabular} & \begin{tabular}[c]{@{}c@{}}96.9 \\ 93.3\end{tabular} & \begin{tabular}[c]{@{}c@{}}96.6 \\ 92.5\end{tabular} & \begin{tabular}[c]{@{}c@{}}97.2 \\ 94.0\end{tabular} & \begin{tabular}[c]{@{}c@{}}96.6 \\ 59.5\end{tabular} & \begin{tabular}[c]{@{}c@{}}60.3 \\ 51.0\end{tabular} & \begin{tabular}[c]{@{}c@{}}95.9 \\ 84.5 \end{tabular} & \begin{tabular}[c]{@{}c@{}}95.6 \\ 50.7\end{tabular} \\ \hline

\multirow{2}{*}{ModelScope} & SigLIP-base & \begin{tabular}[c]{@{}c@{}}F1-Real\\ F1-Fake\end{tabular} & \begin{tabular}[c]{@{}c@{}}99.3 \\ 98.6\end{tabular} & \begin{tabular}[c]{@{}c@{}} - \\ - \end{tabular} & \begin{tabular}[c]{@{}c@{}}99.6 \\ 99.2\end{tabular} & \begin{tabular}[c]{@{}c@{}}99.1 \\ 98.1\end{tabular} & \begin{tabular}[c]{@{}c@{}}99.5 \\ 98.9\end{tabular} & \begin{tabular}[c]{@{}c@{}}95.2 \\ 26.3\end{tabular} & \begin{tabular}[c]{@{}c@{}}56.8 \\ 38.3\end{tabular} & \begin{tabular}[c]{@{}c@{}}96.7 \\ 88.0 \end{tabular} & \begin{tabular}[c]{@{}c@{}}96.1 \\ 58.0\end{tabular} \\ \cline{2-12} 
& VideoMAE & \begin{tabular}[c]{@{}c@{}}F1-Real\\ F1-Fake\end{tabular} & \begin{tabular}[c]{@{}c@{}}96.5 \\ 92.2\end{tabular} & \begin{tabular}[c]{@{}c@{}} - \\ -\end{tabular} & \begin{tabular}[c]{@{}c@{}}95.0 \\ 88.2\end{tabular} & \begin{tabular}[c]{@{}c@{}}95.9 \\ 90.7\end{tabular} & \begin{tabular}[c]{@{}c@{}}95.6 \\ 89.8\end{tabular} & \begin{tabular}[c]{@{}c@{}}95.4 \\ 32.6\end{tabular} & \begin{tabular}[c]{@{}c@{}}54.9 \\ 30.2\end{tabular} & \begin{tabular}[c]{@{}c@{}}94.9 \\ 79.4 \end{tabular} & \begin{tabular}[c]{@{}c@{}}94.9 \\ 36.8\end{tabular} \\ \hline

\multirow{2}{*}{OpenSora} & SigLIP-base & \begin{tabular}[c]{@{}c@{}}F1-Real\\ F1-Fake\end{tabular} & \begin{tabular}[c]{@{}c@{}}99.6 \\ 99.2\end{tabular} & \begin{tabular}[c]{@{}c@{}}97.9 \\ 95.5\end{tabular} & \begin{tabular}[c]{@{}c@{}} - \\ - \end{tabular} & \begin{tabular}[c]{@{}c@{}}97.6\\ 94.8\end{tabular} & \begin{tabular}[c]{@{}c@{}}99.5 \\ 99.0\end{tabular} & \begin{tabular}[c]{@{}c@{}}95.3\\ 28.8\end{tabular} & \begin{tabular}[c]{@{}c@{}}56.8 \\ 38.0\end{tabular} & \begin{tabular}[c]{@{}c@{}}97.6 \\ 91.4\end{tabular} & \begin{tabular}[c]{@{}c@{}}96.5 \\ 63.7\end{tabular} \\ \cline{2-12} 
& VideoMAE & \begin{tabular}[c]{@{}c@{}} F1-Real\\ F1-Fake\end{tabular} & \begin{tabular}[c]{@{}c@{}}97.7 \\ 95.2\end{tabular} & \begin{tabular}[c]{@{}c@{}}97.6 \\ 94.8\end{tabular} & \begin{tabular}[c]{@{}c@{}} - \\ -\end{tabular} & \begin{tabular}[c]{@{}c@{}}97.2 \\ 94.0\end{tabular} & \begin{tabular}[c]{@{}c@{}}96.6 \\ 92.6\end{tabular} & \begin{tabular}[c]{@{}c@{}}96.7 \\ 61.2\end{tabular} & \begin{tabular}[c]{@{}c@{}}63.1 \\ 58.8\end{tabular} & \begin{tabular}[c]{@{}c@{}}96.4 \\ 86.9\end{tabular} & \begin{tabular}[c]{@{}c@{}}96.3 \\ 62.3\end{tabular} \\ \hline

\multirow{2}{*}{ZeroScope} & SigLIP-base & \begin{tabular}[c]{@{}c@{}}F1-Real\\ F1-Fake\end{tabular} & \begin{tabular}[c]{@{}c@{}}99.5 \\ 99.0\end{tabular} & \begin{tabular}[c]{@{}c@{}}99.6 \\ 99.1\end{tabular} & \begin{tabular}[c]{@{}c@{}}99.6 \\ 99.1\end{tabular} & \begin{tabular}[c]{@{}c@{}} - \\ - \end{tabular} & \begin{tabular}[c]{@{}c@{}}99.5 \\ 99.0\end{tabular} & \begin{tabular}[c]{@{}c@{}}96.2 \\ 50.8\end{tabular} & \begin{tabular}[c]{@{}c@{}}63.2 \\ 58.6\end{tabular} & \begin{tabular}[c]{@{}c@{}}97.6 \\ 91.4 \end{tabular} & \begin{tabular}[c]{@{}c@{}}96.7 \\ 66.7\end{tabular} \\ \cline{2-12} 
& VideoMAE & \begin{tabular}[c]{@{}c@{}}F1-Real\\ F1-Fake\end{tabular} & \begin{tabular}[c]{@{}c@{}}96.0 \\ 91.1\end{tabular} & \begin{tabular}[c]{@{}c@{}}97.8 \\ 95.4\end{tabular} & \begin{tabular}[c]{@{}c@{}}95.8 \\ 90.4\end{tabular} & \begin{tabular}[c]{@{}c@{}} - \\ -\end{tabular} & \begin{tabular}[c]{@{}c@{}}95.5 \\ 89.6\end{tabular} & \begin{tabular}[c]{@{}c@{}}96.4 \\ 55.4\end{tabular} & \begin{tabular}[c]{@{}c@{}}59.8 \\ 49.3\end{tabular} & \begin{tabular}[c]{@{}c@{}}95.1 \\ 80.8\end{tabular} & \begin{tabular}[c]{@{}c@{}}95.8 \\ 54.2\end{tabular} \\ \hline

\multirow{2}{*}{Text2Video}  & SigLIP-base & \begin{tabular}[c]{@{}c@{}}F1-Real\\ F1-Fake\end{tabular} & \begin{tabular}[c]{@{}c@{}}99.6 \\ 99.1\end{tabular} & \begin{tabular}[c]{@{}c@{}}98.1 \\ 96.1\end{tabular} & \begin{tabular}[c]{@{}c@{}}99.7 \\ 99.4\end{tabular} & \begin{tabular}[c]{@{}c@{}}97.8 \\ 95.3\end{tabular} & \begin{tabular}[c]{@{}c@{}} - \\ - \end{tabular} & \begin{tabular}[c]{@{}c@{}}95.7\\ 41.3\end{tabular} & \begin{tabular}[c]{@{}c@{}}57.5 \\ 41.3\end{tabular} & \begin{tabular}[c]{@{}c@{}}96.3 \\ 86.2 \end{tabular} & \begin{tabular}[c]{@{}c@{}}96.0 \\ 56.6\end{tabular} \\ \cline{2-12} 
& VideoMAE & \begin{tabular}[c]{@{}c@{}}F1-Real\\ F1-Fake\end{tabular} & \begin{tabular}[c]{@{}c@{}}95.5 \\ 89.7\end{tabular} & \begin{tabular}[c]{@{}c@{}}95.9 \\ 90.7\end{tabular} & \begin{tabular}[c]{@{}c@{}}93.4 \\ 83.7\end{tabular} & \begin{tabular}[c]{@{}c@{}}94.3 \\ 86.4\end{tabular} & \begin{tabular}[c]{@{}c@{}} - \\ -\end{tabular} & \begin{tabular}[c]{@{}c@{}}94.9 \\ 21.4\end{tabular} & \begin{tabular}[c]{@{}c@{}}54.4 \\ 27.7\end{tabular} & \begin{tabular}[c]{@{}c@{}}94.3 \\ 76.3 \end{tabular} & \begin{tabular}[c]{@{}c@{}}95.1 \\40.9\end{tabular} \\ \hline

\multirow{2}{*}{Veo} & SigLIP-base & \begin{tabular}[c]{@{}c@{}}F1-Real\\ F1-Fake\end{tabular} & \begin{tabular}[c]{@{}c@{}}85.0 \\ 45.3\end{tabular} & \begin{tabular}[c]{@{}c@{}}82.2 \\ 23.0\end{tabular} & \begin{tabular}[c]{@{}c@{}}87.1 \\ 58.0\end{tabular} & \begin{tabular}[c]{@{}c@{}}83.3 \\ 33.3\end{tabular} & \begin{tabular}[c]{@{}c@{}}87.6 \\ 60.5\end{tabular} & \begin{tabular}[c]{@{}c@{}} - \\ - \end{tabular} & \begin{tabular}[c]{@{}c@{}}68.0 \\ 69.1\end{tabular} & \begin{tabular}[c]{@{}c@{}}92.2 \\ 63.0 \end{tabular} & \begin{tabular}[c]{@{}c@{}}97.0 \\ 71.1\end{tabular} \\ \cline{2-12} 
& VideoMAE & \begin{tabular}[c]{@{}c@{}}F1-Real\\ F1-Fake\end{tabular} & \begin{tabular}[c]{@{}c@{}}93.5 \\ 83.9\end{tabular} & \begin{tabular}[c]{@{}c@{}}94.6 \\ 87.1\end{tabular} & \begin{tabular}[c]{@{}c@{}}94.3 \\ 86.2\end{tabular} & \begin{tabular}[c]{@{}c@{}}93.8 \\ 84.9\end{tabular} & \begin{tabular}[c]{@{}c@{}}93.9 \\ 85.0\end{tabular} & \begin{tabular}[c]{@{}c@{}} - \\ -\end{tabular} & \begin{tabular}[c]{@{}c@{}}67.1 \\ 67.5\end{tabular} & \begin{tabular}[c]{@{}c@{}}95.0 \\ 79.9 \end{tabular} & \begin{tabular}[c]{@{}c@{}}96.2 \\ 60.1\end{tabular} \\ \hline

\multirow{2}{*}{Sora} & SigLIP-base & \begin{tabular}[c]{@{}c@{}}F1-Real\\ F1-Fake\end{tabular} & \begin{tabular}[c]{@{}c@{}}93.7 \\ 84.5\end{tabular} & \begin{tabular}[c]{@{}c@{}}89.6 \\ 70.1\end{tabular} & \begin{tabular}[c]{@{}c@{}}95.3 \\ 89.2\end{tabular} & \begin{tabular}[c]{@{}c@{}}92.7 \\ 81.5\end{tabular} & \begin{tabular}[c]{@{}c@{}}94.7 \\ 97.5\end{tabular} & \begin{tabular}[c]{@{}c@{}}99.2 \\ 93.0\end{tabular} & \begin{tabular}[c]{@{}c@{}} - \\ - \end{tabular} & \begin{tabular}[c]{@{}c@{}}96.2 \\ 86.1 \end{tabular} & \begin{tabular}[c]{@{}c@{}}98.9 \\ 91.2\end{tabular} \\ \cline{2-12} 
& VideoMAE & \begin{tabular}[c]{@{}c@{}}F1-Real\\ F1-Fake\end{tabular} & \begin{tabular}[c]{@{}c@{}}94.2 \\ 86.1\end{tabular} & \begin{tabular}[c]{@{}c@{}}95.6 \\ 89.9\end{tabular} & \begin{tabular}[c]{@{}c@{}}94.7 \\ 87.4\end{tabular} & \begin{tabular}[c]{@{}c@{}}94.6 \\ 87.3\end{tabular} & \begin{tabular}[c]{@{}c@{}}93.7 \\ 84.5\end{tabular} & \begin{tabular}[c]{@{}c@{}}99.1 \\ 92.0\end{tabular} & \begin{tabular}[c]{@{}c@{}} - \\ - \end{tabular} & \begin{tabular}[c]{@{}c@{}}96.0 \\ 85.0 \end{tabular} & \begin{tabular}[c]{@{}c@{}}97.7 \\ 79.9\end{tabular} \\ \hline

\multirow{2}{*}{Dream Machine} & SigLIP-base & \begin{tabular}[c]{@{}c@{}}F1-Real\\ F1-Fake\end{tabular} & \begin{tabular}[c]{@{}c@{}}94.1 \\ 85.7\end{tabular} & \begin{tabular}[c]{@{}c@{}}90.0 \\ 71.2\end{tabular} & \begin{tabular}[c]{@{}c@{}}95.7 \\ 90.0\end{tabular} & \begin{tabular}[c]{@{}c@{}}88.9 \\ 66.5\end{tabular} & \begin{tabular}[c]{@{}c@{}}94.5 \\ 86.7\end{tabular} & \begin{tabular}[c]{@{}c@{}}97.0 \\ 64.8\end{tabular} & \begin{tabular}[c]{@{}c@{}}58.4 \\ 44.3\end{tabular} & \begin{tabular}[c]{@{}c@{}} - \\ - \end{tabular} & \begin{tabular}[c]{@{}c@{}}97.0 \\ 71.1\end{tabular} \\ \cline{2-12} 
& VideoMAE & \begin{tabular}[c]{@{}c@{}}F1-Real\\ F1-Fake\end{tabular} & \begin{tabular}[c]{@{}c@{}}93.6 \\ 84.2\end{tabular} & \begin{tabular}[c]{@{}c@{}}93.2 \\ 83.1 \end{tabular} & \begin{tabular}[c]{@{}c@{}}93.1 \\ 82.9\end{tabular} & \begin{tabular}[c]{@{}c@{}}91.6 \\ 77.7\end{tabular} & \begin{tabular}[c]{@{}c@{}}92.3 \\ 80.1\end{tabular} & \begin{tabular}[c]{@{}c@{}}97.4 \\ 71.4 \end{tabular} & \begin{tabular}[c]{@{}c@{}} 61.4 \\ 54.2 \end{tabular} & \begin{tabular}[c]{@{}c@{}} - \\ -\end{tabular} & \begin{tabular}[c]{@{}c@{}} 95.6 \\ 51.1 \end{tabular} \\ \hline

\multirow{2}{*}{Video Poet} & SigLIP-base & \begin{tabular}[c]{@{}c@{}}F1-Real\\ F1-Fake\end{tabular} & \begin{tabular}[c]{@{}c@{}}84.5 \\ 41.6\end{tabular} & \begin{tabular}[c]{@{}c@{}}82.7 \\ 27.6\end{tabular} & \begin{tabular}[c]{@{}c@{}}86.3 \\ 53.2\end{tabular} & \begin{tabular}[c]{@{}c@{}}82.2 \\ 23.6\end{tabular} & \begin{tabular}[c]{@{}c@{}}84.9 \\ 44.6\end{tabular} & \begin{tabular}[c]{@{}c@{}}97.5 \\ 73.1\end{tabular} & \begin{tabular}[c]{@{}c@{}}60.3 \\ 50.6\end{tabular} & \begin{tabular}[c]{@{}c@{}}90.3 \\ 48.7 \end{tabular} & \begin{tabular}[c]{@{}c@{}}- \\ -\end{tabular} \\ \cline{2-12} 
& VideoMAE & \begin{tabular}[c]{@{}c@{}}F1-Real\\ F1-Fake\end{tabular} & \begin{tabular}[c]{@{}c@{}}83.0 \\ 30.4\end{tabular} & \begin{tabular}[c]{@{}c@{}}83.2 \\ 31.9\end{tabular} & \begin{tabular}[c]{@{}c@{}}82.7 \\ 28.2\end{tabular} & \begin{tabular}[c]{@{}c@{}}82.5 \\ 25.9\end{tabular} & \begin{tabular}[c]{@{}c@{}}82.5 \\ 26.4\end{tabular} & \begin{tabular}[c]{@{}c@{}}97.7 \\ 75.0\end{tabular} & \begin{tabular}[c]{@{}c@{}}57.9 \\ 42.6\end{tabular} & \begin{tabular}[c]{@{}c@{}}88.5 \\ 29.9 \end{tabular} & \begin{tabular}[c]{@{}c@{}}- \\ -\end{tabular} \\ \hline

\end{tabular}
}
\caption{Evaluation of one-to-many generalization, using training-free, distance-based approach.}
\label{tab:distance_one_to_many}
\end{table*}

\begin{table*}[h!]
\resizebox{\linewidth}{!}{
\begin{tabular}{cccccccccccc}
\hline

\multirow{2}{*}{Reference} & \multirow{2}{*}{Model} & \multirow{2}{*}{Metric} & \multicolumn{7}{c}{Testing (real and fake)} \\ \cline{4-12}
& & & Latte & ModelScope & OpenSora & ZeroScope & Text2Video & Veo & Sora & Dream Machine & Video Poet \\ \hline

\multirow{2}{*}{Open-source models} & SigLIP-base & \begin{tabular}[c]{@{}c@{}}F1-Real\\ F1-Fake\end{tabular} & \begin{tabular}[c]{@{}c@{}}99.8\\ 99.2\end{tabular} & \begin{tabular}[c]{@{}c@{}}99.8 \\ 99.1\end{tabular} & \begin{tabular}[c]{@{}c@{}}99.8 \\ 99.0\end{tabular} & \begin{tabular}[c]{@{}c@{}}99.7 \\ 98.8\end{tabular} & \begin{tabular}[c]{@{}c@{}}99.8 \\ 99.1\end{tabular} & \begin{tabular}[c]{@{}c@{}}96.2\\ 51.1\end{tabular} & \begin{tabular}[c]{@{}c@{}}95.4\\ 70.3\end{tabular} & \begin{tabular}[c]{@{}c@{}}98.1\\ 93.5\end{tabular} & \begin{tabular}[c]{@{}c@{}}97.3 \\ 74.3\end{tabular} \\ \cline{2-12}
& VideoMAE & \begin{tabular}[c]{@{}c@{}}F1-Real\\ F1-Fake\end{tabular} & \begin{tabular}[c]{@{}c@{}}98.8 \\ 94.9\end{tabular} & \begin{tabular}[c]{@{}c@{}}99.2 \\ 96.5\end{tabular} & \begin{tabular}[c]{@{}c@{}}98.8 \\ 95.0\end{tabular} & \begin{tabular}[c]{@{}c@{}}98.9 \\ 95.3\end{tabular} & \begin{tabular}[c]{@{}c@{}}99.2 \\ 96.8\end{tabular} & \begin{tabular}[c]{@{}c@{}}97.1 \\ 67.0\end{tabular} & \begin{tabular}[c]{@{}c@{}}94.6 \\ 55.3\end{tabular} & \begin{tabular}[c]{@{}c@{}}96.8\\ 88.3\end{tabular} & \begin{tabular}[c]{@{}c@{}}96.5 \\ 64.0\end{tabular} \\ \hline

\multirow{2}{*}{Sora} & SigLIP-base & \begin{tabular}[c]{@{}c@{}}F1-Real\\ F1-Fake\end{tabular} & \begin{tabular}[c]{@{}c@{}}95.9 \\ 79.8\end{tabular} & \begin{tabular}[c]{@{}c@{}}93.7 \\ 64.2\end{tabular} & \begin{tabular}[c]{@{}c@{}}97.3 \\ 87.6\end{tabular} & \begin{tabular}[c]{@{}c@{}}95.8 \\ 79.2\end{tabular} & \begin{tabular}[c]{@{}c@{}}96.4 \\ 82.9\end{tabular} & \begin{tabular}[c]{@{}c@{}}99.0 \\ 91.1\end{tabular} & \begin{tabular}[c]{@{}c@{}} - \\ - \end{tabular} & \begin{tabular}[c]{@{}c@{}}95.6\\ 83.0\end{tabular} & \begin{tabular}[c]{@{}c@{}}98.9  \\ 91.2\end{tabular} \\ \cline{2-12}
& VideoMAE & \begin{tabular}[c]{@{}c@{}}F1-Real\\ F1-Fake\end{tabular} & \begin{tabular}[c]{@{}c@{}}96.8 \\ 84.7\end{tabular} & \begin{tabular}[c]{@{}c@{}}97.5 \\ 88.8\end{tabular} & \begin{tabular}[c]{@{}c@{}}97.1 \\ 86.4\end{tabular} & \begin{tabular}[c]{@{}c@{}}97.0 \\ 86.2\end{tabular} & \begin{tabular}[c]{@{}c@{}}96.5 \\ 83.3\end{tabular} & \begin{tabular}[c]{@{}c@{}}99.1 \\ 92.0\end{tabular} & \begin{tabular}[c]{@{}c@{}} -\\ -\end{tabular} & \begin{tabular}[c]{@{}c@{}}96.0\\85.0\end{tabular} & \begin{tabular}[c]{@{}c@{}}97.7 \\ 79.9\end{tabular} \\ \hline

\multirow{2}{*}{Open-source models + Sora} & SigLIP-base & \begin{tabular}[c]{@{}c@{}}F1-Real\\ F1-Fake\end{tabular} & \begin{tabular}[c]{@{}c@{}}99.8 \\ 99.2\end{tabular} & \begin{tabular}[c]{@{}c@{}}99.8 \\ 99.1\end{tabular} & \begin{tabular}[c]{@{}c@{}}99.8 \\ 99.0\end{tabular} & \begin{tabular}[c]{@{}c@{}}99.7 \\ 98.8\end{tabular} & \begin{tabular}[c]{@{}c@{}}99.8 \\ 99.1\end{tabular} & \begin{tabular}[c]{@{}c@{}}97.3 \\ 70.2\end{tabular} & \begin{tabular}[c]{@{}c@{}}99.3 \\ 96.6\end{tabular} & \begin{tabular}[c]{@{}c@{}}98.1\\ 93.6\end{tabular} & \begin{tabular}[c]{@{}c@{}}97.8 \\ 80.5\end{tabular} \\ \cline{2-12}
& VideoMAE & \begin{tabular}[c]{@{}c@{}}F1-Real\\ F1-Fake\end{tabular} & \begin{tabular}[c]{@{}c@{}}98.8 \\ 94.9\end{tabular} & \begin{tabular}[c]{@{}c@{}}99.2 \\ 96.5\end{tabular} & \begin{tabular}[c]{@{}c@{}}98.8 \\ 95.0\end{tabular} & \begin{tabular}[c]{@{}c@{}}99.0 \\ 95.7\end{tabular} & \begin{tabular}[c]{@{}c@{}}99.3 \\ 96.9\end{tabular} & \begin{tabular}[c]{@{}c@{}}97.1 \\ 67.0\end{tabular} & \begin{tabular}[c]{@{}c@{}}98.6 \\ 91.5\end{tabular} & \begin{tabular}[c]{@{}c@{}}97.2\\ 89.9\end{tabular} & \begin{tabular}[c]{@{}c@{}}96.6 \\ 66.0\end{tabular} \\ \hline

\end{tabular}
}
\caption{Evaluation of many-to-many generalization, using training-free, distance-based approach.}
\label{tab:distance_many_to_many}
\vspace{-0.4cm}
\end{table*}

\begin{table*}[h!]
\resizebox{\linewidth}{!}{
\begin{tabular}{cccccccccccc}
\hline
\multirow{2}{*}{Training} & \multirow{2}{*}{Model} & \multirow{2}{*}{Metric} & \multicolumn{7}{c}{Testing (real and fake)} \\ \cline{4-12}
& & & Latte & ModelScope & OpenSora & ZeroScope & Text2Video & Veo & Sora & Dream Machine & Video Poet\\ \hline

\multirow{2}{*}{Latte} & SigLIP-base & \begin{tabular}[c]{@{}c@{}}F1-Real\\ F1-Fake\end{tabular} & \begin{tabular}[c]{@{}c@{}} -\\ -\end{tabular} & \begin{tabular}[c]{@{}c@{}}98.9 \\ 98.9\end{tabular} & \begin{tabular}[c]{@{}c@{}}100.0 \\ 99.9\end{tabular} & \begin{tabular}[c]{@{}c@{}}99.2 \\ 99.2\end{tabular} & \begin{tabular}[c]{@{}c@{}}100.0 \\ 100.0\end{tabular} & \begin{tabular}[c]{@{}c@{}}90.8 \\ 25.6\end{tabular} & \begin{tabular}[c]{@{}c@{}}36.3 \\ 21.2 \end{tabular}  & \begin{tabular}[c]{@{}c@{}}98.5 \\ 97.6\end{tabular} & \begin{tabular}[c]{@{}c@{}}99.1 \\ 96.6\end{tabular} \\ \cline{2-12} 
& VideoMAE & \begin{tabular}[c]{@{}c@{}}F1-Real\\ F1-Fake\end{tabular} & \begin{tabular}[c]{@{}c@{}} -\\ -\end{tabular} & \begin{tabular}[c]{@{}c@{}}97.2 \\ 97.0\end{tabular} & \begin{tabular}[c]{@{}c@{}}87.7 \\ 84.0\end{tabular} & \begin{tabular}[c]{@{}c@{}}95.6 \\ 95.2\end{tabular} & \begin{tabular}[c]{@{}c@{}}98.4 \\ 98.3\end{tabular} & \begin{tabular}[c]{@{}c@{}}91.3 \\ 38.0\end{tabular} & \begin{tabular}[c]{@{}c@{}}38.8 \\ 35.8\end{tabular} & \begin{tabular}[c]{@{}c@{}}90.9 \\ 82.0\end{tabular} & \begin{tabular}[c]{@{}c@{}}91.5 \\ 53.1\end{tabular} \\ \hline

\multirow{2}{*}{ModelScope} & SigLIP-base & \begin{tabular}[c]{@{}c@{}}F1-Real\\ F1-Fake\end{tabular} & \begin{tabular}[c]{@{}c@{}}100.0 \\ 99.9\end{tabular} & \begin{tabular}[c]{@{}c@{}} -\\ -\end{tabular} & \begin{tabular}[c]{@{}c@{}}99.9 \\ 99.8\end{tabular} & \begin{tabular}[c]{@{}c@{}}100.0 \\ 100.0\end{tabular} & \begin{tabular}[c]{@{}c@{}}100.0 \\ 100.0\end{tabular} & \begin{tabular}[c]{@{}c@{}}90.1 \\ 13.3\end{tabular} & \begin{tabular}[c]{@{}c@{}}35.4 \\ 15.3\end{tabular}  & \begin{tabular}[c]{@{}c@{}}94.3 \\ 89.5\end{tabular} & \begin{tabular}[c]{@{}c@{}}98.9 \\ 95.8\end{tabular} \\ \cline{2-12} 
& VideoMAE & \begin{tabular}[c]{@{}c@{}}F1-Real\\ F1-Fake\end{tabular} & \begin{tabular}[c]{@{}c@{}}91.3 \\ 89.5\end{tabular} & \begin{tabular}[c]{@{}c@{}} -\\ -\end{tabular} & \begin{tabular}[c]{@{}c@{}}76.9 \\ 57.6\end{tabular} & \begin{tabular}[c]{@{}c@{}}91.9 \\ 90.5\end{tabular} & \begin{tabular}[c]{@{}c@{}}92.1 \\ 90.7\end{tabular} & \begin{tabular}[c]{@{}c@{}}89.9 \\ 14.6\end{tabular} & \begin{tabular}[c]{@{}c@{}}35.1 \\ 14.1\end{tabular}  & \begin{tabular}[c]{@{}c@{}}83.0 \\ 53.1\end{tabular} & \begin{tabular}[c]{@{}c@{}}88.5 \\ 11.6\end{tabular} \\ \hline

\multirow{2}{*}{OpenSora} & SigLIP-base & \begin{tabular}[c]{@{}c@{}}F1-Real\\ F1-Fake\end{tabular} & \begin{tabular}[c]{@{}c@{}}100.0 \\ 100.0\end{tabular} & \begin{tabular}[c]{@{}c@{}}98.7 \\ 98.7\end{tabular} & \begin{tabular}[c]{@{}c@{}} -\\ -\end{tabular} & \begin{tabular}[c]{@{}c@{}}99.4\\ 99.3\end{tabular} & \begin{tabular}[c]{@{}c@{}}100.0 \\ 100.0\end{tabular} & \begin{tabular}[c]{@{}c@{}}90.8 \\ 25.0\end{tabular} & \begin{tabular}[c]{@{}c@{}}36.9 \\ 24.8\end{tabular}  & \begin{tabular}[c]{@{}c@{}}98.4 \\ 97.4\end{tabular} & \begin{tabular}[c]{@{}c@{}}99.2 \\ 97.0\end{tabular} \\ \cline{2-12} 
& VideoMAE & \begin{tabular}[c]{@{}c@{}} F1-Real\\ F1-Fake\end{tabular} & \begin{tabular}[c]{@{}c@{}}94.1 \\ 93.7\end{tabular} & \begin{tabular}[c]{@{}c@{}}92.7 \\ 91.9\end{tabular} & \begin{tabular}[c]{@{}c@{}} -\\ -\end{tabular} & \begin{tabular}[c]{@{}c@{}}94.4 \\ 94.0\end{tabular} & \begin{tabular}[c]{@{}c@{}}94.8 \\ 94.5\end{tabular} & \begin{tabular}[c]{@{}c@{}}94.6 \\ 73.0\end{tabular} & \begin{tabular}[c]{@{}c@{}}48.2 \\ 64.4\end{tabular}  & \begin{tabular}[c]{@{}c@{}}92.4 \\ 86.0\end{tabular} & \begin{tabular}[c]{@{}c@{}}93.8 \\ 72.3\end{tabular} \\ \hline

\multirow{2}{*}{ZeroScope} & SigLIP-base & \begin{tabular}[c]{@{}c@{}}F1-Real\\ F1-Fake\end{tabular} & \begin{tabular}[c]{@{}c@{}}99.9 \\ 99.9\end{tabular} & \begin{tabular}[c]{@{}c@{}}99.6 \\ 99.6\end{tabular} & \begin{tabular}[c]{@{}c@{}}99.9 \\ 99.8\end{tabular} & \begin{tabular}[c]{@{}c@{}} -\\ -\end{tabular} & \begin{tabular}[c]{@{}c@{}}100.0 \\ 100.0\end{tabular} & \begin{tabular}[c]{@{}c@{}}89.9 \\ 11.1\end{tabular} & \begin{tabular}[c]{@{}c@{}}37.0 \\ 25.1\end{tabular}  & \begin{tabular}[c]{@{}c@{}}94.6 \\ 90.1\end{tabular} & \begin{tabular}[c]{@{}c@{}}98.9 \\ 95.6\end{tabular} \\ \cline{2-12} 
& VideoMAE & \begin{tabular}[c]{@{}c@{}}F1-Real\\ F1-Fake\end{tabular} & \begin{tabular}[c]{@{}c@{}}92.8 \\ 91.7\end{tabular} & \begin{tabular}[c]{@{}c@{}}96.2 \\ 95.9\end{tabular} & \begin{tabular}[c]{@{}c@{}}87.2 \\ 83.0\end{tabular} & \begin{tabular}[c]{@{}c@{}} -\\ -\end{tabular} & \begin{tabular}[c]{@{}c@{}}95.9 \\ 95.6\end{tabular} & \begin{tabular}[c]{@{}c@{}}91.2 \\ 33.8\end{tabular} & \begin{tabular}[c]{@{}c@{}}41.8 \\ 46.6\end{tabular}  & \begin{tabular}[c]{@{}c@{}}86.6 \\ 68.3\end{tabular} & \begin{tabular}[c]{@{}c@{}}91.6 \\ 51.5\end{tabular} \\ \hline

\multirow{2}{*}{Text2Video}  & SigLIP-base & \begin{tabular}[c]{@{}c@{}}F1-Real\\ F1-Fake\end{tabular} & \begin{tabular}[c]{@{}c@{}}99.2 \\ 99.1\end{tabular} & \begin{tabular}[c]{@{}c@{}}93.4 \\ 92.4\end{tabular} & \begin{tabular}[c]{@{}c@{}}99.4 \\ 99.4\end{tabular} & \begin{tabular}[c]{@{}c@{}}94.7 \\ 94.1\end{tabular} & \begin{tabular}[c]{@{}c@{}} -\\ -\end{tabular} & \begin{tabular}[c]{@{}c@{}}90.8 \\ 26.3\end{tabular} & \begin{tabular}[c]{@{}c@{}}35.8 \\ 17.8\end{tabular}  & \begin{tabular}[c]{@{}c@{}}89.1 \\ 75.8\end{tabular} & \begin{tabular}[c]{@{}c@{}}93.2 \\ 63.3\end{tabular} \\ \cline{2-12} 
& VideoMAE & \begin{tabular}[c]{@{}c@{}}F1-Real\\ F1-Fake\end{tabular} & \begin{tabular}[c]{@{}c@{}}82.7 \\ 73.8\end{tabular} & \begin{tabular}[c]{@{}c@{}}84.6 \\ 78.0\end{tabular} & \begin{tabular}[c]{@{}c@{}}71.1 \\ 32.1\end{tabular} & \begin{tabular}[c]{@{}c@{}}80.7 \\ 68.7\end{tabular} & \begin{tabular}[c]{@{}c@{}} -\\ -\end{tabular} & \begin{tabular}[c]{@{}c@{}}89.6 \\ 8.0\end{tabular} & \begin{tabular}[c]{@{}c@{}}35.0 \\ 13.6\end{tabular}  & \begin{tabular}[c]{@{}c@{}}79.0 \\ 28.0\end{tabular} & \begin{tabular}[c]{@{}c@{}}89.5 \\ 25.5\end{tabular} \\ \hline

\multirow{2}{*}{Veo} & SigLIP-base & \begin{tabular}[c]{@{}c@{}}F1-Real\\ F1-Fake\end{tabular} & \begin{tabular}[c]{@{}c@{}}76.1 \\ 54.4\end{tabular} & \begin{tabular}[c]{@{}c@{}}68.6 \\ 16.5\end{tabular} & \begin{tabular}[c]{@{}c@{}}79.7 \\ 66.1\end{tabular} & \begin{tabular}[c]{@{}c@{}}71.7 \\ 35.4\end{tabular} & \begin{tabular}[c]{@{}c@{}}85.1 \\ 78.9\end{tabular} & \begin{tabular}[c]{@{}c@{}} -\\ -\end{tabular} & \begin{tabular}[c]{@{}c@{}}56.4 \\ 76.1\end{tabular}  & \begin{tabular}[c]{@{}c@{}}85.8 \\ 64.5\end{tabular} & \begin{tabular}[c]{@{}c@{}}96.7 \\ 85.8\end{tabular} \\ \cline{2-12} 
& VideoMAE & \begin{tabular}[c]{@{}c@{}}F1-Real\\ F1-Fake\end{tabular} & \begin{tabular}[c]{@{}c@{}}70.1 \\ 26.6\end{tabular} & \begin{tabular}[c]{@{}c@{}}72.2 \\ 37.9\end{tabular} & \begin{tabular}[c]{@{}c@{}}71.7 \\ 35.4\end{tabular} & \begin{tabular}[c]{@{}c@{}}70.7 \\ 30.0\end{tabular} & \begin{tabular}[c]{@{}c@{}}70.1 \\ 26.3\end{tabular} & \begin{tabular}[c]{@{}c@{}} -\\ -\end{tabular} & \begin{tabular}[c]{@{}c@{}}47.4 \\ 61.6\end{tabular}  & \begin{tabular}[c]{@{}c@{}}81.9 \\ 46.5\end{tabular} & \begin{tabular}[c]{@{}c@{}}94.3 \\ 72.0\end{tabular} \\ \hline

\multirow{2}{*}{Sora} & SigLIP-base & \begin{tabular}[c]{@{}c@{}}F1-Real\\ F1-Fake\end{tabular} & \begin{tabular}[c]{@{}c@{}}94.0 \\ 93.3\end{tabular} & \begin{tabular}[c]{@{}c@{}}87.1 \\ 82.9\end{tabular} & \begin{tabular}[c]{@{}c@{}}96.2 \\ 96.0\end{tabular} & \begin{tabular}[c]{@{}c@{}}94.7 \\ 94.1\end{tabular} & \begin{tabular}[c]{@{}c@{}}96.0 \\ 95.8\end{tabular} & \begin{tabular}[c]{@{}c@{}}99.2 \\ 96.6\end{tabular} & \begin{tabular}[c]{@{}c@{}} -\\ -\end{tabular}  & \begin{tabular}[c]{@{}c@{}}95.8 \\ 92.8\end{tabular} & \begin{tabular}[c]{@{}c@{}}99.6 \\ 98.7\end{tabular} \\ \cline{2-12} 
& VideoMAE & \begin{tabular}[c]{@{}c@{}}F1-Real\\ F1-Fake\end{tabular} & \begin{tabular}[c]{@{}c@{}}76.4 \\ 61.4\end{tabular} & \begin{tabular}[c]{@{}c@{}}74.4 \\ 55.3\end{tabular} & \begin{tabular}[c]{@{}c@{}}83.0 \\ 77.3\end{tabular} & \begin{tabular}[c]{@{}c@{}}83.7 \\ 78.8\end{tabular} & \begin{tabular}[c]{@{}c@{}}80.0 \\ 71.0\end{tabular} & \begin{tabular}[c]{@{}c@{}}96.0 \\ 84.8\end{tabular} & \begin{tabular}[c]{@{}c@{}} -\\ -\end{tabular}  & \begin{tabular}[c]{@{}c@{}}87.3 \\ 75.9\end{tabular} & \begin{tabular}[c]{@{}c@{}}94.1 \\ 79.5\end{tabular} \\ \hline

\multirow{2}{*}{Dream Machine} & SigLIP-base & \begin{tabular}[c]{@{}c@{}}F1-Real\\ F1-Fake\end{tabular} & \begin{tabular}[c]{@{}c@{}}97.9 \\ 97.8\end{tabular} & \begin{tabular}[c]{@{}c@{}}88.7 \\ 85.4\end{tabular} & \begin{tabular}[c]{@{}c@{}}98.0 \\ 98.0\end{tabular} & \begin{tabular}[c]{@{}c@{}}84.5 \\ 77.6\end{tabular} & \begin{tabular}[c]{@{}c@{}}97.3 \\ 97.2\end{tabular} & \begin{tabular}[c]{@{}c@{}}90.7 \\ 23.7\end{tabular} & \begin{tabular}[c]{@{}c@{}} 37.6\\ 28.8\end{tabular}  & \begin{tabular}[c]{@{}c@{}}- \\ -\end{tabular} & \begin{tabular}[c]{@{}c@{}}99.6 \\ 98.7\end{tabular} \\ \cline{2-12} 
& VideoMAE & \begin{tabular}[c]{@{}c@{}}F1-Real\\ F1-Fake\end{tabular} & \begin{tabular}[c]{@{}c@{}}92.5 \\ 91.4\end{tabular} & \begin{tabular}[c]{@{}c@{}}86.5\\ 82.2\end{tabular} & \begin{tabular}[c]{@{}c@{}}90.1 \\ 88.1\end{tabular} & \begin{tabular}[c]{@{}c@{}}88.8 \\ 86.0\end{tabular} & \begin{tabular}[c]{@{}c@{}}91.9 \\ 90.7\end{tabular} & \begin{tabular}[c]{@{}c@{}}96.1 \\ 81.5\end{tabular} & \begin{tabular}[c]{@{}c@{}} 53.2\\72.3 \end{tabular}  & \begin{tabular}[c]{@{}c@{}} -\\ -\end{tabular} & \begin{tabular}[c]{@{}c@{}}94.8 \\ 76.4\end{tabular} \\ \hline

\multirow{2}{*}{Video Poet} & SigLIP-base & \begin{tabular}[c]{@{}c@{}}F1-Real\\ F1-Fake\end{tabular} & \begin{tabular}[c]{@{}c@{}}79.3 \\ 64.8\end{tabular} & \begin{tabular}[c]{@{}c@{}}72.7 \\ 39.6\end{tabular} & \begin{tabular}[c]{@{}c@{}}85.3 \\ 79.2\end{tabular} & \begin{tabular}[c]{@{}c@{}}69.5 \\ 21.9\end{tabular} & \begin{tabular}[c]{@{}c@{}}74.8 \\ 49.2\end{tabular} & \begin{tabular}[c]{@{}c@{}}89.8 \\ 8.8\end{tabular} & \begin{tabular}[c]{@{}c@{}} 36.0\\ 19.2\end{tabular}  & \begin{tabular}[c]{@{}c@{}}83.2 \\ 52.9\end{tabular} & \begin{tabular}[c]{@{}c@{}}- \\ -\end{tabular} \\ \cline{2-12} 
& VideoMAE & \begin{tabular}[c]{@{}c@{}}F1-Real\\ F1-Fake\end{tabular} & \begin{tabular}[c]{@{}c@{}}73.6 \\ 45.0\end{tabular} & \begin{tabular}[c]{@{}c@{}}71.6\\ 35.4\end{tabular} & \begin{tabular}[c]{@{}c@{}}76.3 \\ 55.6\end{tabular} & \begin{tabular}[c]{@{}c@{}}78.5 \\ 62.9\end{tabular} & \begin{tabular}[c]{@{}c@{}}78.2 \\ 61.9\end{tabular} & \begin{tabular}[c]{@{}c@{}}96.7 \\ 84.0\end{tabular} & \begin{tabular}[c]{@{}c@{}} 50.0\\66.8\end{tabular}  & \begin{tabular}[c]{@{}c@{}} 83.7\\ 56.3\end{tabular} & \begin{tabular}[c]{@{}c@{}}- \\ -\end{tabular} \\ \hline

\end{tabular}
}
\caption{Evaluation of one-to-many generalization, using training-based approach.}
\label{tab:train_one_to_many}
\end{table*}

\begin{table*}[h!]
\resizebox{\linewidth}{!}{
\begin{tabular}{cccccccccccc}
\hline

\multirow{2}{*}{Training} & \multirow{2}{*}{Model} & \multirow{2}{*}{Metric} & \multicolumn{7}{c}{Testing (real and fake)} \\ \cline{4-12}
& & & Latte & ModelScope & OpenSora & ZeroScope & Text2Video & Veo & Sora & Dream Machine & Video Poet\\ \hline

\multirow{2}{*}{Open-source models} & SigLIP-base & \begin{tabular}[c]{@{}c@{}}F1-Real\\ F1-Fake\end{tabular} & \begin{tabular}[c]{@{}c@{}}100.0\\ 100.0\end{tabular} & \begin{tabular}[c]{@{}c@{}}100.0 \\ 99.9\end{tabular} & \begin{tabular}[c]{@{}c@{}}100.0 \\ 100.0\end{tabular} & \begin{tabular}[c]{@{}c@{}}100.0 \\ 100.0\end{tabular} & \begin{tabular}[c]{@{}c@{}}100.0 \\ 100.0\end{tabular} & \begin{tabular}[c]{@{}c@{}}95.2\\ 25.0\end{tabular} & \begin{tabular}[c]{@{}c@{}}91.3\\ 16.2\end{tabular} & \begin{tabular}[c]{@{}c@{}}99.4\\ 98.0\end{tabular} & \begin{tabular}[c]{@{}c@{}}99.8 \\ 98.7\end{tabular} \\ \cline{2-12}
& VideoMAE & \begin{tabular}[c]{@{}c@{}}F1-Real\\ F1-Fake\end{tabular} & \begin{tabular}[c]{@{}c@{}}99.1 \\ 96.4\end{tabular} & \begin{tabular}[c]{@{}c@{}}99.5 \\ 97.9\end{tabular} & \begin{tabular}[c]{@{}c@{}}99.0 \\ 96.1\end{tabular} & \begin{tabular}[c]{@{}c@{}}99.3 \\ 97.4\end{tabular} & \begin{tabular}[c]{@{}c@{}}99.5 \\ 98.2\end{tabular} & \begin{tabular}[c]{@{}c@{}}96.5 \\ 60.3\end{tabular} & \begin{tabular}[c]{@{}c@{}}96.6 \\ 78.2\end{tabular} & \begin{tabular}[c]{@{}c@{}}96.9\\ 89.0\end{tabular} & \begin{tabular}[c]{@{}c@{}}97.1 \\ 74.2\end{tabular} \\ \hline

\multirow{2}{*}{Sora} & SigLIP-base & \begin{tabular}[c]{@{}c@{}}F1-Real\\ F1-Fake\end{tabular} & \begin{tabular}[c]{@{}c@{}}98.6 \\ 94.0\end{tabular} & \begin{tabular}[c]{@{}c@{}}97.0 \\ 86.2\end{tabular} & \begin{tabular}[c]{@{}c@{}}99.2 \\ 96.6\end{tabular} & \begin{tabular}[c]{@{}c@{}}98.3 \\ 92.5\end{tabular} & \begin{tabular}[c]{@{}c@{}}98.8 \\ 95.0\end{tabular} & \begin{tabular}[c]{@{}c@{}}99.3 \\ 94.1\end{tabular} & \begin{tabular}[c]{@{}c@{}}- \\ -\end{tabular} & \begin{tabular}[c]{@{}c@{}}92.7\\ 71.6\end{tabular} & \begin{tabular}[c]{@{}c@{}}99.9 \\ 99.1\end{tabular} \\ \cline{2-12}
& VideoMAE & \begin{tabular}[c]{@{}c@{}}F1-Real\\ F1-Fake\end{tabular} & \begin{tabular}[c]{@{}c@{}}91.7 \\ 54.1\end{tabular} & \begin{tabular}[c]{@{}c@{}}91.1 \\ 48.6\end{tabular} & \begin{tabular}[c]{@{}c@{}}93.6 \\ 68.9\end{tabular} & \begin{tabular}[c]{@{}c@{}}94.0 \\ 71.4\end{tabular} & \begin{tabular}[c]{@{}c@{}}93.0 \\ 64.6\end{tabular} & \begin{tabular}[c]{@{}c@{}}97.6 \\ 80.6\end{tabular} & \begin{tabular}[c]{@{}c@{}}- \\ -\end{tabular} & \begin{tabular}[c]{@{}c@{}}92.7\\ 71.6\end{tabular} & \begin{tabular}[c]{@{}c@{}}97.2 \\ 79.1\end{tabular} \\ \hline

\multirow{2}{*}{Open-source models + Sora} & SigLIP-base & \begin{tabular}[c]{@{}c@{}}F1-Real\\ F1-Fake\end{tabular} & \begin{tabular}[c]{@{}c@{}}98.9 \\ 95.8\end{tabular} & \begin{tabular}[c]{@{}c@{}}99.2 \\ 97.0\end{tabular} & \begin{tabular}[c]{@{}c@{}}98.9 \\ 95.4\end{tabular} & \begin{tabular}[c]{@{}c@{}}99.1 \\ 96.5\end{tabular} & \begin{tabular}[c]{@{}c@{}}99.2 \\ 97.1\end{tabular} & \begin{tabular}[c]{@{}c@{}}97.0 \\ 70.3\end{tabular} & \begin{tabular}[c]{@{}c@{}}98.9 \\ 93.8\end{tabular} & \begin{tabular}[c]{@{}c@{}}97.4\\ 91.4\end{tabular} & \begin{tabular}[c]{@{}c@{}}97.6 \\ 80.3\end{tabular} \\ \cline{2-12}
& VideoMAE & \begin{tabular}[c]{@{}c@{}}F1-Real\\ F1-Fake\end{tabular} & \begin{tabular}[c]{@{}c@{}}98.9 \\ 95.8\end{tabular} & \begin{tabular}[c]{@{}c@{}}99.2 \\ 96.9\end{tabular} & \begin{tabular}[c]{@{}c@{}}98.9 \\ 95.4\end{tabular} & \begin{tabular}[c]{@{}c@{}}99.1 \\ 96.5\end{tabular} & \begin{tabular}[c]{@{}c@{}}99.2 \\ 97.1\end{tabular} & \begin{tabular}[c]{@{}c@{}}97.0 \\ 70.3\end{tabular} & \begin{tabular}[c]{@{}c@{}}98.9 \\ 93.8\end{tabular} & \begin{tabular}[c]{@{}c@{}}97.4\\ 91.4\end{tabular} & \begin{tabular}[c]{@{}c@{}}97.5 \\ 80.3\end{tabular} \\ \hline

\end{tabular}
}
\caption{Evaluation of many-to-many generalization, using training-based approach.}
\label{tab:train_many_to_many}
\vspace{-0.2cm}
\end{table*}

With the training-free approach, we achieve better results in the many-to-many protocol where we use videos from all the models in the reference set. 
In the one-to-many setting, we achieve the best performance across the different generation models, when using Sora \cite{videoworldsimulators2024} videos as reference. Overall, as expected, we observed better performance using our training-based approach with both one-to-many and many-to-many protocols. 

Also, from these results, we observe that our models achieve better detection performance on the open-source 
models, compared to the latest state-of-the-art closed-source T2V generation models. This highlights the advanced capabilities of the latest T2V generation models, which produce more realistic videos of much higher quality compared to open-source models, making detection significantly more challenging. Please refer to supplementary material for a detailed discussion of these results. 
\section{Conclusion}
In this work, we address the challenge of detecting AI-generated videos, which extends beyond the scope of traditional Deepfake detection. While Deepfake detection focuses primarily on videos of humans, our approach is designed to handle videos with more diverse and generic content. To address this problem, we propose training-based and training-free approaches that utilize video features extracted from pre-trained visual models. These models, having been trained on large-scale real-world video datasets, capture the distribution of real content. Hence, the features from these models contain the signals needed to distinguish between real and AI-generated videos.

Given the lack of publicly available datasets for this task, we curated a dataset comprising over 7 hours of video data, including more than 10,000 videos from 9 different T2V generation models and 4,000 real videos for evaluation. Our dataset contains videos with diverse content and includes videos from multiple open-source T2V models along with high-quality videos generated by the latest closed-source models. These characteristics of our dataset make it a very valuable resource for the development and evaluation of detection methods for AI-generated videos. 

Our experimental results demonstrate strong detection performance across both open-source and closed-source models. However, we observe reduced accuracy when detecting videos from the latest closed-source models, highlighting the advancements in generative technology. We hope that our dataset and findings will inspire further research in this area, as detecting AI-generated videos is vital for combating the spread of misinformation and fake news.
\section*{Supplement}
The following is an overview of the content in this section.
\begin{itemize}
    \item In Section \ref{comparison}, we show the results of our model on the recently released GenVideo \cite{DeMamba} dataset.
    \item In Section \ref{dataset}, we present additional details about our dataset, including some sample captions used to generate videos in the dataset using the open-source text-to-video (T2V) models. 
    \item In Section \ref{future_work}, we discuss the possible future work, with the extension of our method to detect spatial and temporal inconsistencies in the videos generated by the T2V models.
\end{itemize}

\section{Ours vs GenVideo} \label{comparison}

Developing and evaluating methods for detecting AI-generated videos requires a dataset with diverse content from multiple video generation models. At the start of our work, no such dataset was publicly available, leading us to create our own to address this gap. However, the recent release of GenVideo \cite{DeMamba}, a large-scale dataset featuring videos generated by various AI models, also serves the same purpose. This dataset contains over 2 million training and nearly 20,000 testing videos, equally balanced between real and fake videos. It offers diverse, high-quality content with fake videos generated using multiple models, covering a wide range of scenes and resolutions. Compared to GenVideo, our proposed dataset contains videos from the most recent generation models such as Veo \cite{veo}, Dream Machine \cite{dreammachine}, and Video Poet \cite{kondratyukvideopoet}.

In \citet{DeMamba}, the authors propose an approach involving training a state-space model (DeMamba) on their large-scale GenVideo dataset to detect AI-generated videos. Their model contains 125.37M trainable parameters and is trained on the dataset containing 2M generated videos. In contrast, our approach involves training a single linear layer with only 1.5K parameters on features extracted from pre-trained visual models, using just 4.5K generated videos (many-to-many protocol). For a fair comparison, we evaluate our models on the GenVideo test set following the many-to-many protocol and present the results in Table \ref{tab:many_to_many}. Similarly, we also present a comparison of results following the one-to-many protocol in Table \ref{tab:one_to_many}. Here, for comparison, we show results with our model trained on OpenSora videos to ensure consistency in the evaluation. 

These results show that our method achieves better performance than DeMamba \cite{DeMamba} on both protocols, using a much smaller dataset for training and a simple, parameter-efficient model with a minimal number of trainable parameters. Specifically, our method achieves 7.9\% improvement over DeMamba \cite{DeMamba} using the one-to-many protocol and 0.9\% improvement over a comparable model with the many-to-many protocol. These results of our method are with the image-based SigLIP model, whereas DeMamba achieves its best performance using a video-based XCLIP model. Using features from a video-based XCLIP model with our approach would increase the gains of our model even further. 

\begin{table*}[h!]
\resizebox{\linewidth}{!}{
\begin{tabular}{c|c|ccccccccccc|cc}
\hline

\multirow{2}{*}{Model } & \multirow{2}{*}{Detection Level} & \multicolumn{8}{c}{Testing (real and fake)} \\ \cline{3-14}
& & Sora & Gen2 & MorphStudio & ModelScope & Show & Lavie & Wildscrape & Crafter & MoonValley & HotShot & Real & Average\\ \hline

\multirow{1}{*}{CLIP-B-PT* (2M)} & Image & \begin{tabular}[c]{@{}c@{}} 85.7 \end{tabular} & \begin{tabular}[c]{@{}c@{}} 90.4 \end{tabular} & \begin{tabular}[c]{@{}c@{}} 82.4 \end{tabular} & \begin{tabular}[c]{@{}c@{}} 82.1 \end{tabular} & \begin{tabular}[c]{@{}c@{}} 75.4 \end{tabular} & \begin{tabular}[c]{@{}c@{}} 79.3 \end{tabular} & \begin{tabular}[c]{@{}c@{}} 75.2 \end{tabular} & \begin{tabular}[c]{@{}c@{}} 86.3  \end{tabular} & \begin{tabular}[c]{@{}c@{}} 89.6 \end{tabular} 
 & \begin{tabular}[c]{@{}c@{}} 71.0 \end{tabular}  & \begin{tabular}[c]{@{}c@{}} 57.2 \end{tabular}  & \begin{tabular}[c]{@{}c@{}} 79.7 \end{tabular} \\ \cline{2-14} 
\hline

\multirow{1}{*}{DeMamba-CLIP-B-FT* (2M)} & Video & \begin{tabular}[c]{@{}c@{}} 95.7 \end{tabular} & \begin{tabular}[c]{@{}c@{}} 100.0 \end{tabular} & \begin{tabular}[c]{@{}c@{}} 98.7 \end{tabular} & \begin{tabular}[c]{@{}c@{}} 69.1 \end{tabular} & \begin{tabular}[c]{@{}c@{}} 92.4 \end{tabular} & \begin{tabular}[c]{@{}c@{}} 93.2 \end{tabular} & \begin{tabular}[c]{@{}c@{}} 100.0 \end{tabular} & \begin{tabular}[c]{@{}c@{}} 100.0  \end{tabular} & \begin{tabular}[c]{@{}c@{}} 83.6 \end{tabular} 
 & \begin{tabular}[c]{@{}c@{}} 82.9 \end{tabular}  & \begin{tabular}[c]{@{}c@{}} 99.4 \end{tabular}  & \begin{tabular}[c]{@{}c@{}} 92.3 \end{tabular} \\ \cline{2-14} 
\hline
\hline
\rowcolor{gray}
\multirow{1}{*}{SigLIP-base (4.5K) } & Image & \begin{tabular}[c]{@{}c@{}} 98.2 \end{tabular} & \begin{tabular}[c]{@{}c@{}} 99.4 \end{tabular} & \begin{tabular}[c]{@{}c@{}} 95.9 \end{tabular} & \begin{tabular}[c]{@{}c@{}} 91.6 \end{tabular} & \begin{tabular}[c]{@{}c@{}} 94.9 \end{tabular} & \begin{tabular}[c]{@{}c@{}} 97.3 \end{tabular} & \begin{tabular}[c]{@{}c@{}} 98.7 \end{tabular} & \begin{tabular}[c]{@{}c@{}} 98.6  \end{tabular} & \begin{tabular}[c]{@{}c@{}} 99.0 \end{tabular} 
 & \begin{tabular}[c]{@{}c@{}} 95.3 \end{tabular}  & \begin{tabular}[c]{@{}c@{}} 56.8 \end{tabular}  & \begin{tabular}[c]{@{}c@{}} \textbf{93.2 (+0.9)} \end{tabular} \\ \cline{2-14} 
\hline

 \end{tabular}
}
\caption{Comparison of our results with DeMamba \cite{DeMamba}. The results shown in this table are the accuracy score with the many-to-many protocol. Our results are shown in the row, colored gray. Results indicated with * are from \citet{DeMamba}. The values in parentheses show the size of the data set used to train the model.}
\label{tab:many_to_many}
\end{table*}

\begin{table*}[ht!]
\resizebox{\linewidth}{!}{
\begin{tabular}{c|c|ccccccccccc|cc}
\hline

\multirow{2}{*}{Training Set (Size)} & \multirow{2}{*}{Model} & \multicolumn{8}{c}{Testing (real and fake)} \\ \cline{3-14}
& & Sora & Gen2 & MorphStudio & ModelScope & Show & Lavie & Wildscrape & Crafter & MoonValley & HotShot & Real & Average \\ \hline

\multirow{1}{*}{OpenSora (177K)} & NPR* & \begin{tabular}[c]{@{}c@{}} 55.4 \end{tabular} & \begin{tabular}[c]{@{}c@{}} 55.5 \end{tabular} & \begin{tabular}[c]{@{}c@{}} 76.3 \end{tabular} & \begin{tabular}[c]{@{}c@{}} 29.9 \end{tabular} & \begin{tabular}[c]{@{}c@{}} 22.4 \end{tabular} & \begin{tabular}[c]{@{}c@{}} 76.5 \end{tabular} & \begin{tabular}[c]{@{}c@{}} 60.4 \end{tabular} & \begin{tabular}[c]{@{}c@{}} 83.1  \end{tabular} & \begin{tabular}[c]{@{}c@{}} 74.9 \end{tabular} 
 & \begin{tabular}[c]{@{}c@{}} 58.6 \end{tabular}  & \begin{tabular}[c]{@{}c@{}} 95.9 \end{tabular}  & \begin{tabular}[c]{@{}c@{}} 62.6 \end{tabular} \\ \cline{2-14} 
& DeMamba-XCLIP-FT*  & \begin{tabular}[c]{@{}c@{}} 55.4 \end{tabular} & \begin{tabular}[c]{@{}c@{}} 81.3 \end{tabular} & \begin{tabular}[c]{@{}c@{}} 87.4 \end{tabular} & \begin{tabular}[c]{@{}c@{}} 44.9 \end{tabular} & \begin{tabular}[c]{@{}c@{}} 73.1 \end{tabular} & \begin{tabular}[c]{@{}c@{}} 85.2 \end{tabular} & \begin{tabular}[c]{@{}c@{}} 58.1 \end{tabular} & \begin{tabular}[c]{@{}c@{}} 90.1 \end{tabular} & \begin{tabular}[c]{@{}c@{}} 89.6 \end{tabular}  & \begin{tabular}[c]{@{}c@{}} 73.1 \end{tabular}  & \begin{tabular}[c]{@{}c@{}} 97.3 \end{tabular} 
 & \begin{tabular}[c]{@{}c@{}} 75.9 \end{tabular} \\ 
 \hline
 \hline
\rowcolor{gray}
\multirow{1}{*}{OpenSora (1K) } & SigLIP-base & \begin{tabular}[c]{@{}c@{}} 87.5 \end{tabular} & \begin{tabular}[c]{@{}c@{}} 95.3 \end{tabular} & \begin{tabular}[c]{@{}c@{}} 71.1 \end{tabular} & \begin{tabular}[c]{@{}c@{}} 54.7 \end{tabular} & \begin{tabular}[c]{@{}c@{}} 77.0 \end{tabular} & \begin{tabular}[c]{@{}c@{}} 89.7 \end{tabular} & \begin{tabular}[c]{@{}c@{}} 93.6 \end{tabular} & \begin{tabular}[c]{@{}c@{}} 93.4  \end{tabular} & \begin{tabular}[c]{@{}c@{}} 92.6 \end{tabular} 
 & \begin{tabular}[c]{@{}c@{}} 82.6 \end{tabular}  & \begin{tabular}[c]{@{}c@{}} 83.8 \end{tabular}  & \begin{tabular}[c]{@{}c@{}} \textbf{83.8 (+7.9) } \end{tabular} \\ \cline{2-14} 
\hline

\end{tabular}
}
\caption{Comparison of our results with DeMamba \cite{DeMamba}. The results shown in this table are the accuracy score with the one-to-many protocol. Our results are shown in the row, colored in gray. Results indicated with * are from \citet{DeMamba}.}
\label{tab:one_to_many}
\end{table*}

\section{Dataset} \label{dataset}

Here are few example captions we used to generate videos in our dataset. These captions themselves are generated by GPT3.5, using the prompt we provided in the main paper. As you can see, these captions describe varied scenes containing various objects and different actors performing different actions.  

\begin{itemize}
    \item A woman in a red dress twirls gracefully on a wooden deck, the sun setting behind her.
    \item A young boy in a blue shirt jumps into a swimming pool, creating a big splash.
    \item A cat with orange fur leaps from a kitchen counter to a dining table, narrowly missing a glass vase.
    \item An elderly man in a brown coat feeds pigeons in a bustling city square, smiling gently.
    \item A dog with a red collar chases a green tennis ball across a grassy park.
    \item A chef in a white hat flips a pancake in a busy restaurant kitchen, catching it perfectly on the pan.
    \item A toddler in a yellow raincoat jumps into a puddle, water splashing up around them.
    \item A cyclist in a blue jersey speeds down a mountain trail, the forest blurred behind them.
    \item A barista with a beard pours steamed milk into a cup, creating a heart-shaped latte art.
    \item A woman in a yoga pose stretches her arms towards the sky on a serene beach at dawn.
\end{itemize}

\begin{figure*}[h!]
  \centering
  \includegraphics[width=0.75\linewidth]{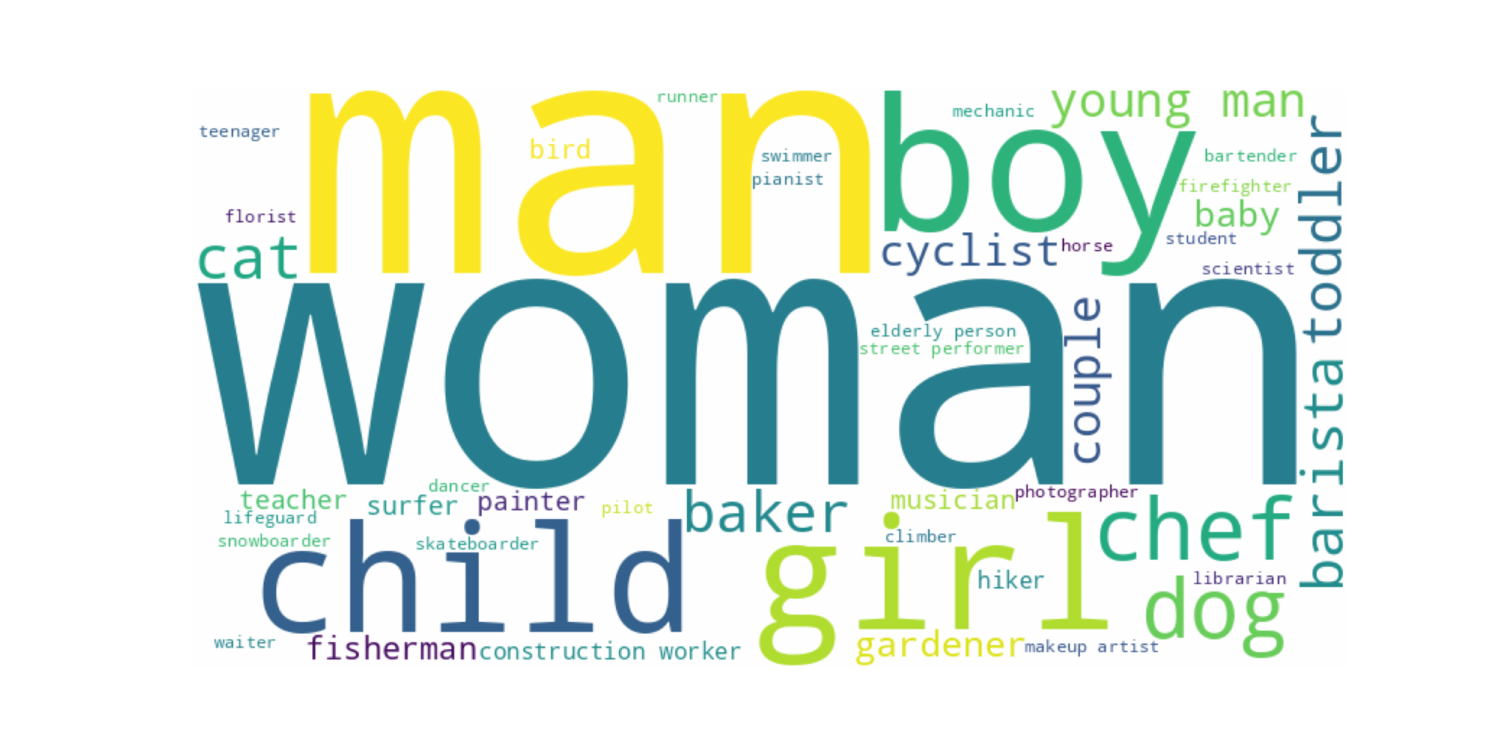}
   \caption{Word Cloud showing the diversity of actors in the videos from our dataset.}
   \label{fig:subjects}
   \vspace{-0.4cm}
\end{figure*}

\begin{figure*}[h!]
  \centering
  \includegraphics[width=0.75\linewidth]{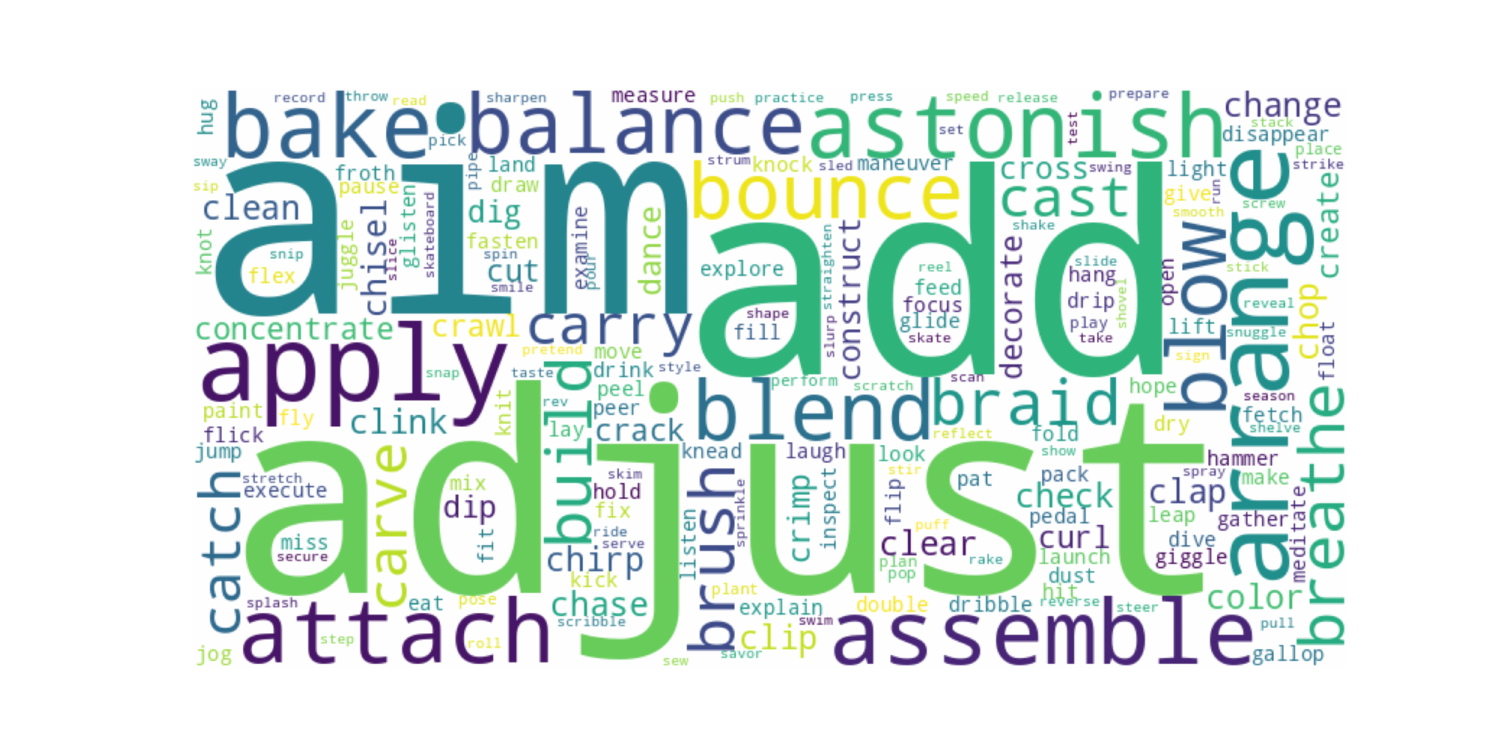}
   \caption{Word Cloud showing all the verbs, corresponding to the actions depicted in the videos in our dataset.}
   \label{fig:verbs}
   \vspace{-0.4cm}
\end{figure*}

\begin{figure*}[h!]
  \centering
  \includegraphics[width=0.75\linewidth]{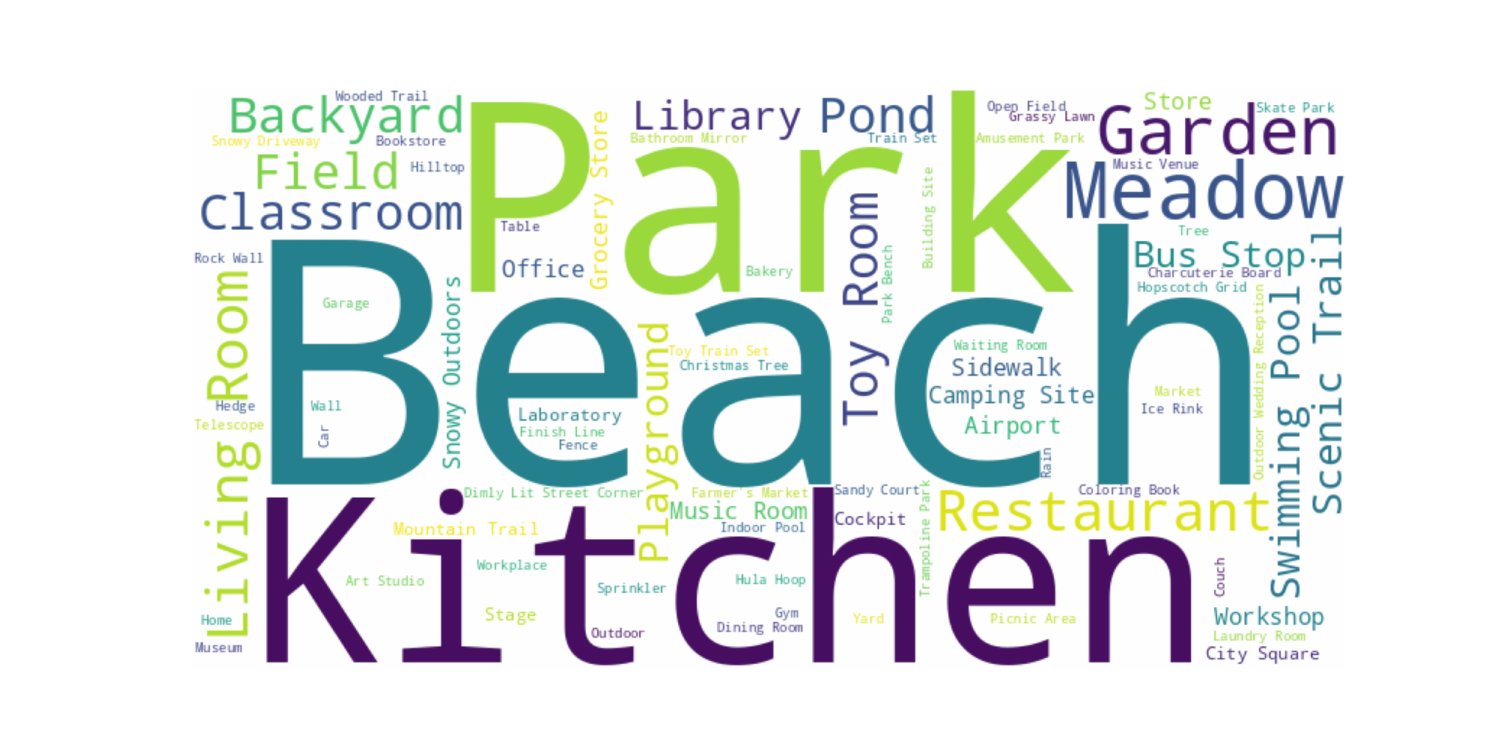}
   \caption{Word Cloud showing different scenes captured in the videos from our dataset.}
   \label{fig:scenes}
\end{figure*}

Please refer to Figure \ref{fig:subjects}, showing the different types of actors, including animals along with humans. In addition, humans of both genders of different age groups of different professions are included. This shows one aspect of diversity in our dataset. Next in Figure \ref{fig:verbs}, we show the word cloud showing $221$ distinct verbs from the captions. These show the diversity in our dataset compared to the actions depicted in the videos from our dataset. Finally, in Figure \ref{fig:scenes}, we show the scenes described in the captions used to generate the videos in our dataset. This shows that the videos in our dataset are from both indoors and outdoors, covering different locations, adding to the diversity of our dataset.

\section{Future Work} \label{future_work}
In our experiments, we have used only the features of the SigLIP (image) and VideoMAE (video) visual models. One possible future work is to explore the use of other visual models such as XCLIP \cite{ni2022expanding} or V-JEPA \cite{bardes2023v} to improve our performance. The other possible direction for future work is to extend our method to learn to detect the inconsistent spatio-temporal patches in the generated video. This provides interpretability of the results by grounding the predictions of the model.

{
    \small
    \bibliographystyle{ieeenat_fullname}
    \bibliography{main}
}


\end{document}